\documentclass[10pt,twocolumn,letterpaper]{article}

\usepackage{cvpr}              

\usepackage[accsupp]{axessibility}

\usepackage[dvipsnames]{xcolor}

\definecolor{cvprblue}{rgb}{0.21,0.49,0.74}
\usepackage[pagebackref,breaklinks,colorlinks,citecolor=cvprblue]{hyperref}

\def\paperID{10991} 
\def\confName{CVPR}
\def\confYear{2024}

\def\eg{\textit{e.g.}}
\def\ie{\textit{i.e.}}

\newcommand{\Skip}[1]{}

\usepackage{multirow}
\usepackage{soul}

\title{UFORecon: Generalizable Sparse-View Surface Reconstruction \\ from Arbitrary and Unfavorable Sets}

\author{Youngju Na \qquad Woo Jae Kim \qquad Kyu Beom Han \qquad Suhyeon Ha \qquad Sung-Eui Yoon\\
KAIST \\
{\tt\small \{yjna2907,wkim97,qbhan,suhyeon.ha,sungeui\}@kaist.ac.kr}
}

\begin{document}
\maketitle
\begin{abstract}

Generalizable neural implicit surface reconstruction aims to obtain an accurate underlying geometry given a limited number of multi-view images from unseen scenes.
However, existing methods select only informative and relevant views using predefined scores for training and testing phases.
This constraint makes the model impractical because we cannot always ensure the availability of favorable combinations in real-world scenarios.
We observe that previous methods output degenerate solutions under arbitrary and unfavorable sets.
Building upon this finding, we propose \textbf{UFORecon}, a robust view-combination generalizable surface reconstruction framework.
To this end, we apply cross-view matching transformers to model interactions between source images and build correlation frustums to capture global correlations.
In addition, we explicitly encode pairwise feature similarities as view-consistent priors. 
Our proposed framework largely outperforms previous methods not only in view-combination generalizability but also in the existing generalizable protocol trained with favorable view-combinations. The code is available at \href{https://github.com/Youngju-Na/UFORecon}{https://github.com/Youngju-Na/UFORecon}.
\end{abstract}    
\section{Introduction}
Reconstructing 3D geometries from images of multiple camera viewpoints is a fundamental problem in the computer vision field, also applicable to robotics~\cite{niceslam,graspnerf,lens}, autonomous driving~\cite{driving,read}, and AR/VR applications~\cite{scalable,deitke2023objaverse}.
Multi-view stereo (MVS) is a widely used technique for reconstructing 3D geometry from multi-view images. Conventional MVS methods~\cite{depthhypotheses, parallelmvs, carving, colmap} find correspondences across input images taken from different viewpoints.
With the recent advances in deep learning, learning-based MVS methods~\cite{mvsnet,casmvsnet,transmvsnet,rc-mvsnset} leverage differentiable homography warping with 3D Convolutional Neural Network (CNN) to effectively search correspondences in feature space. 
\begin{figure}
    \centering 
    \includegraphics[width=\linewidth, trim={0 12.6cm 17.9cm 0cm}, clip]{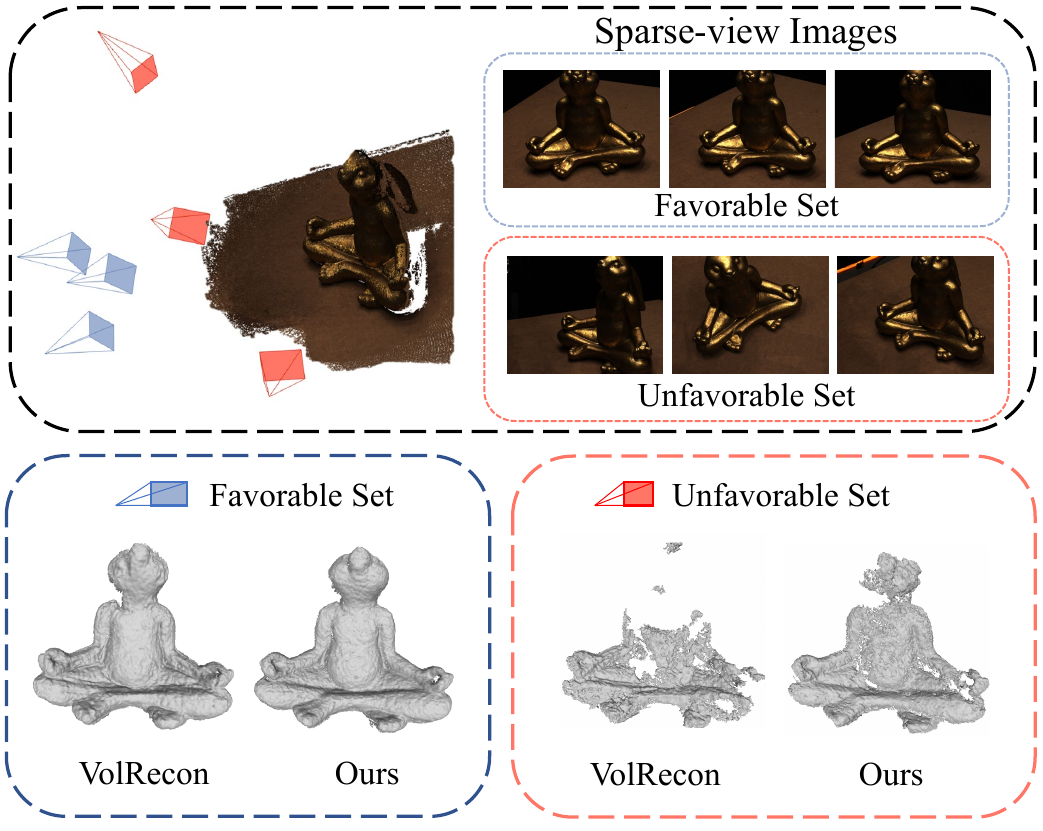} 
    \caption{\textbf{Reconstruction results from different view combinations}. Both are trained only with the best-selected training protocol and tested with favorable (\textcolor{Blue}{Blue}) and unfavorable sets (\textcolor{Red}{Red}), respectively. VolRecon~\cite{volrecon} leads to a degenerate geometry in the unfavorable set while achieving accurate geometry in the favorable set. Our approach produces reasonable geometry on both sets.}
    \label{fig:concept}
    \vspace{-10pt}
\end{figure}
Recent progress in 3D reconstruction~\cite{idr,deepsdf,sdfdiff} increasingly emphasizes the use of implicit neural representation~\cite{nerf}. 
This trend is notable in the field of neural surface reconstruction~\cite{neus, volsdf, unisurf, monosdf}.
Unlike Neural Radiance Fields (NeRF)~\cite{nerf}, which employs a volume density to infer 3D geometry implicitly, the use of Signed Distance Function (SDF) field has proven to be effective for learning surfaces because of their universal zero level-set definition of surfaces.
However, as these approaches optimize on a single scene with dense camera viewpoints, there is an increasing effort to learn a generalizable representation that can adaptively reconstruct geometry from unseen scenes with sparse camera viewpoints~\cite{sparseneus,volrecon,retr,c2f2neus}. 

Despite the remarkable achievements, existing generalizable implicit surface reconstruction methods use the fixed, best combination for each camera based on the view-selection scores~\cite{mvsnet}. This score is predefined for each camera pair via a matching-based reconstruction, such as COLMAP~\cite{colmap}.
These \textit{favorably} selected views usually share a significant amount of overlaps, which can help the model easily capture the 3D geometry of the target scene.
We argue that this assumption limits the practicality of surface reconstruction tasks in two key scenarios: (1) where inferring from arbitrary combinations of camera views that were not available during training and (2) where the combination of views share a limited amount of overlapping areas with each other for reconstruction.


To verify this argument, we start by introducing the view-combination score (\textit{VC} Score) that measures the informativeness of the input image set for reconstruction. 
We demonstrate that conventional research gradually faces higher difficulty as the \textit{VC} score decreases.
We then introduce our proposed framework, UFORecon, a robust view-combination generalizable surface reconstruction framework.
More specifically, we learn the matching features between source view images using an attention-based cross-view transformer.
Then, we build a cascaded correlation frustum based on each source view, effectively combining MVS with implicit surface reconstruction for robust reconstruction from varied view combinations.
Furthermore, we encode the explicit similarity for all input pairs and intermediate depths into sample points to predict a 3D representation. 
As a result, we achieve effective view-combination generalizability without additional data or losses.
Finally, we introduce a random set training strategy for robustness under arbitrary view combinations.

As shown in Figure~\ref{fig:concept}, while existing methods suffer from degraded performance under the unfavorable set, our proposed method achieves a reasonable geometry. We attribute this to our method's capability to learn the correlation among source view images.


\noindent
In summary, our contributions are as follows:
\begin{itemize}
    \setlength{\topsep}{0pt}
    \setlength{\parskip}{0pt}
    \item We propose a new concept of \textit{view-combination generalizability}, which represents an ability to reconstruct geometry under arbitrary and unfavorable views, in generalizable neural scene representation.
    \item We propose an effective model that integrates cross-view features with correlation volumes and explicit feature similarities to learn cross-view interaction and reconstruct under arbitrary and unfavorable sets. 
    \item We validate the efficacy of random set training as a robust strategy to stably enhance view-combination generalizability.
\end{itemize}

\section{Related Works}
\subsection{Depth Map-based Multi-view Stereo}
Multi-view stereo (MVS) is a branch of the 3D reconstruction task that estimates the underlying geometry of a scene or an object given images captured from multiple views. 
The common method for classical MVS is finding correspondence or matching features of multi-view images and projecting them into a certain 3D representation~\cite{sfm79}.
Among various representations (\eg,~point clouds~\cite{stereopsis,quasi}, voxel grids~\cite{lsm,surfacenet}, depths map~\cite{sfm16,depthhypotheses,parallelmvs}), depths map-based MVS has shown to be effective in recent learning-based MVS thanks to their robustness and flexibility.
MVSNet~\cite{mvsnet} proposes to encode camera parameters and deep image features with differentiable homography to build 3D cost volume. 
CasMVSNet~\cite{casmvsnet} further reduces the memory consumption of the cost volume by proposing a coarse-to-fine multi-stage depth estimation framework.
TransMVSNet~\cite{transmvsnet} proposes a feature-matching transformer to extract cross-view correlation and robust global context across multi-view images.
Though these methods have shown promising results, they show limited results in non-Lambertian regions, and their estimation capability is limited to estimating the depths of known images, while volume rendering-based implicit methods could render color or depth images of novel viewpoints. 

\subsection{Implicit Neural Scene Reconstruction}
Implicit neural representations for 3D scenes have gained significant attention since NeRF~\cite{nerf}, which represents the 3D scene as an implicit continuous signal, approximated with neural networks, \eg,~multi-layer perceptrons (MLPs). Especially in surface reconstruction, NeuS~\cite{neus} and VolSDF~\cite{volsdf} exploit implicit neural representations for signed distance fields (SDFs) in 3D scenes.
Utilizing geometric cues~\cite{monosdf} (e.g., depth, normal) or adding constraints~\cite{sparf} has proven effective in sparse-view reconstruction. However, these approaches barely consider the relationship among source views and involve intensive test-time optimization. To address this, generalizable neural scene representation methods construct feature volumes~\cite{mvsnerf,pixelnerf} or aggregate source views using transformer attention modules~\cite{ibrnet,common}.
Particularly for surface reconstruction, SparseNeuS~\cite{sparseneus} constructs hierarchical feature volumes using deep image features to enable geometry-aware reasoning. VolRecon~\cite{volrecon} and ReTR~\cite{retr} utilize transformers to effectively fuse multi-view image features. C2F2NeuS~\cite{c2f2neus} effectively integrates MVS and implicit surface reconstruction by building cascaded frustum to encode global geometry features.
However, these methods' performance is often bounded by camera sets with large overlaps and exhibits sensitivity to viewpoint variations. In our research, we model feature correlation among images and employ view-consistent matching prior to effectively handling arbitrary or unfavorable view sets. 

\subsection{Feature Matching Transformer}
Transformers, which primarily utilize attention mechanisms~\cite{attention}, have been increasingly employed in diverse 3D understanding applications. Notable among these are transformer-based MVS methods introduced by STTR~\cite{sttr}, MVSTER~\cite{mvster}, and TransMVSNet~\cite{transmvsnet}. These methods significantly improve feature matching by employing both self- and cross-attention techniques to effectively train on features from multiple viewpoints.
Other than the MVS methods, GMFlow~\cite{gmflow} exploits a similar transformer-based architecture to estimate optical flow by reformulating the estimation as a global image feature-matching problem between frames. More recently, MatchNeRF~\cite{matchnerf} introduces cross-view feature matching from the pre-trained transformer in GMFlow~\cite{gmflow} as an explicit geometry guidance for generalizable Novel View Synthesis (NVS).
Despite these advancements, the exploration of cross-view feature matching in implicit 3D reconstruction remains underdeveloped. 
Our approach effectively combines transformer-based feature correlation with neural implicit surface methods, significantly enhancing view-combination generalizability and 3D reconstruction. 


\begin{figure}[t]
    \centering
    \includegraphics[width=\linewidth, trim={0cm 0cm 0cm 0cm}, clip]{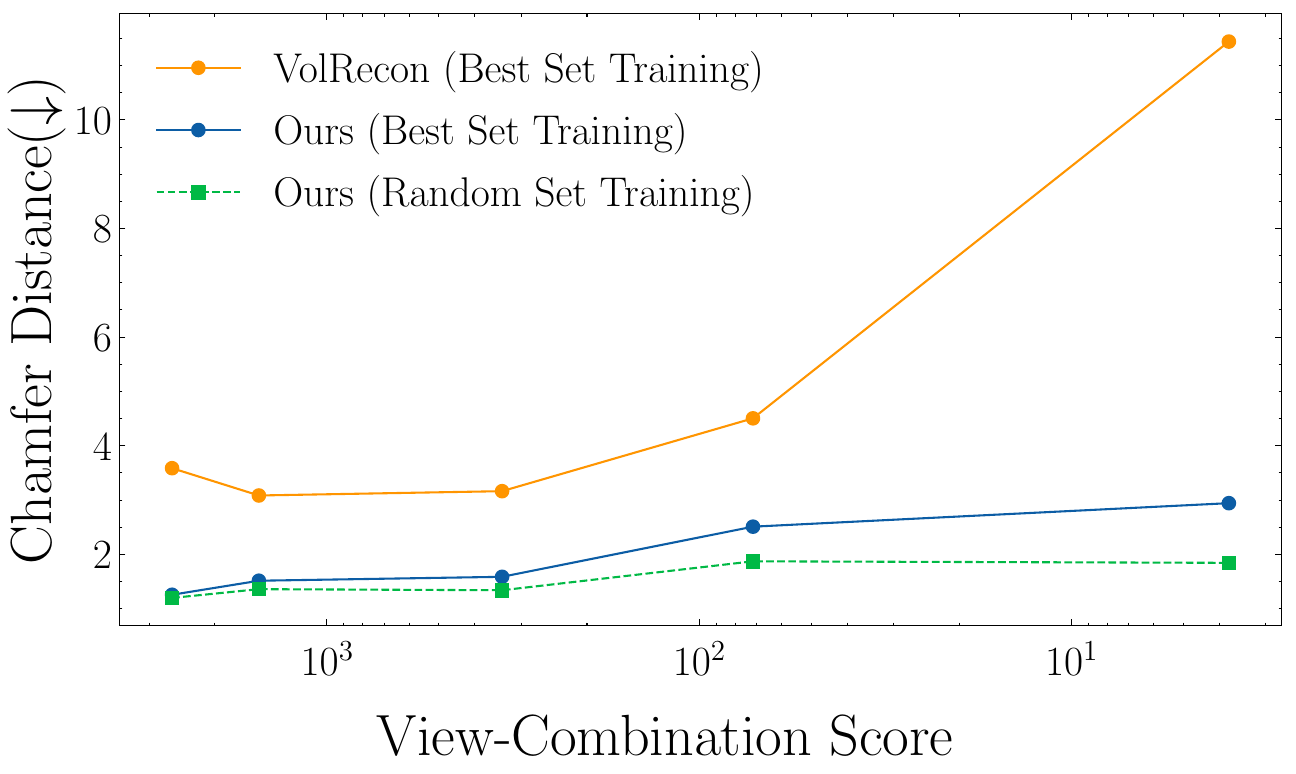}
    \caption{\textbf{Comparison of Chamfer Distance (CD) by view-combination (\textit{VC}) scores for generalizable implicit surface reconstruction methods.} We define the VC score to represent the informativeness of view combinations in reconstruction. The higher VC score represents a more favorable combination. Our method shows better generalizability and accuracy over VolRecon~\cite{volrecon} across all VC scores. Our random set training (Sec.~\ref{Training}) further improves the view-combination generalizability.}
    \label{fig:cd_fav}
\end{figure}

\section{Motivation}\label{sec:motivation}


The existing generalizable methods adhere to two critical assumptions that limit their ability to effectively reconstruct from varying view combinations.
The assumptions are:
1) The network is trained on images with the best view-selection scores for the specific target view.
2) The evaluation similarly assumes the favorable combination for reconstruction by selecting the source cameras with high view selection scores or selecting the neighboring cameras to the target camera.
We define a property of a view-combination that deviates from such assumptions as \textit{unfavorable}.
In contrast, favorable combinations tend to share a relatively large overlap in their camera view frustum, and the respective camera positions are likely to be cluttered (~\ie, small baseline and rotation).
We observe that a network trained only with such combinations easily overfits, restricting generalizability under arbitrary view-combination variations.


To measure the effectiveness of a view-combination for the reconstruction, we define a new metric, \textit{view-combination score} (\textit{VC} score), to evaluate the collective informativeness of a viewpoints combination for 3D reconstruction.
We derive the score from view selection score~\cite{mvsnet}, which represents the usefulness of an image pair for reconstruction.
The view selection score~\cite{mvsnet} $s(i,j)$ between two cameras $i$ and $j$ are defined as follows:
\begin{equation}
s(i,j)=\sum_\textbf{p}(\mathcal{G}(\theta_{ij}(\textbf{p})),
\end{equation}
where $\mathcal{G}$ is a piecewise Gaussian function that favors a certain baseline angle $\theta_{ij}$, which denotes the angle between the two camera centers. \textbf{p} is a common track in both view $i$ and $j$ obtained with COLMAP~\cite{colmap}.
The complete derivation process of the view-selection score and the hyperparameters is thoroughly detailed in the Appendix.
We extend it to a set of multiple images by averaging all of the pairwise scores between images in the set as follows:
\begin{equation}
\textit{VC} = \frac{1}{\binom{n}{2}} \sum_{i=1}^{n-1} \sum_{j=i+1}^{n} s(i,j).
\end{equation}
\Skip{Intuitively, our \textit{VC} score reflects the overall informativeness of the combination for the reconstruction.}
To show the validity of the VC score and verify the conventional approach suffers from the degradation of performance under unfavorable combinations, we test the view-generalizability of the previous approach in various VC score levels.
As depicted in Figure \ref{fig:cd_fav}, Chamfer Distance (CD) incrementally rises as the VC score reduces in the previous state-of-the-art (SoTA) reconstruction method~\cite{volrecon}.
Such result reflects the necessity of providing a prior on the informativeness of image combination for robust reconstruction.
Inspired by these findings, our objective centers on enhancing \textit{view-combination generalizability} by providing cross-view correlation features as robust prior for the image combination, boosting the quality of reconstructions under arbitrary and unfavorable sets of images.




\begin{figure*}[t]
    \centering
    \includegraphics[width=0.9\linewidth, trim={0 12.5cm 1.1cm 0cm}, clip]{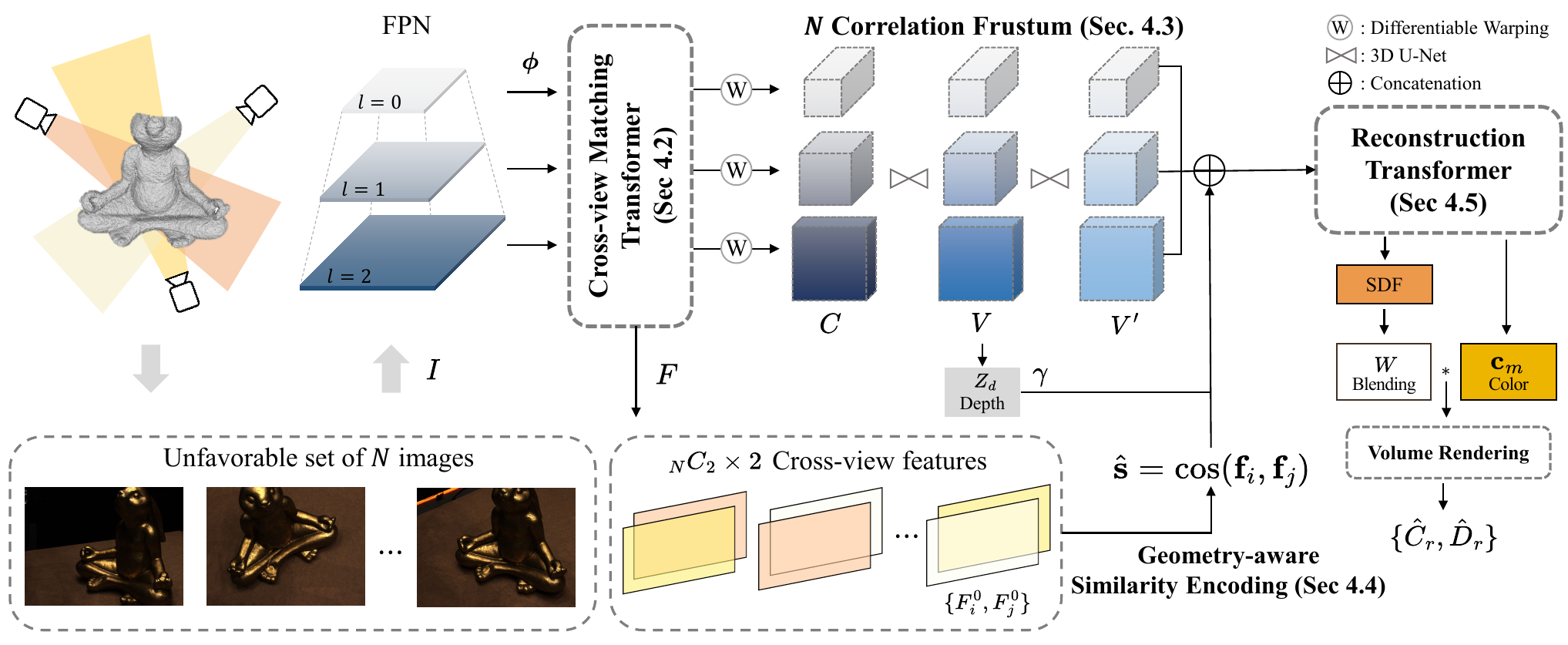}
    \caption{\textbf{Overall pipeline of UFORecon}. Our cross-view matching transformer extracts cross-view matching features from multi-level image features of the given image set (Sec.~\ref{cross-view transformer}). Cross-view matching features are then represented as 3D volumes (\ie,~cascaded correlation frustums, (Sec.~\ref{cross-view volume}) and as the 2D features (\ie,~geometry aware similarity encoding, (Sec.~\ref{similarity encoding}). Our reconstruction transformer (Sec.~\ref{aggregation transformer}) fuses various representations of matching features and geometry features (\ie,~ depth) for volume rendering and color blending.}
    \label{fig:framework}
    \vspace{-0.3cm}
\end{figure*}

\section{Methods}
In this section, we first introduce the general formulations of volume rendering-based generalizable surface reconstruction (Sec.~\ref{preliminaries}). Then, we introduce the overall structure of our proposed UFORecon (Figure~\ref{fig:framework}) which consists of cross-view matching feature extraction (Sec.~\ref{cross-view transformer}), cascaded correlation volume generation strategy (Sec.~\ref{cross-view volume}) that encodes global geometry across input views, explicit 2D geometry-aware matching priors (Sec.~\ref{similarity encoding}) that provides robust view-pair guidance, and view aggregation transformer (Sec.~\ref{aggregation transformer}) to aggregate global volumes and 2D priors with self-attention transformers. Finally, we introduce our training schemes (Sec.~\ref{Training}).
\vspace{-0.1cm}

\subsection{Preliminaries}\label{preliminaries}
The goal of generalizable neural surface reconstruction is to reconstruct the underlying geometry of an arbitrary scene with the posed images \(\{I_i, P_i\}_{i=1}^{N}\) where \( {I_i} \in \mathbb{R}^{H \times W \times 3} \) is the $i$-th view's image and \( P_i \in \mathbb{R}^{3 \times 4} \) is the corresponding camera parameters.
A general framework extracts features from 2D images and builds a global feature volume \cite{sparseneus, volrecon, retr, c2f2neus}. Then the NeRF-like MLP decoder leverages these\Skip{global-local} features with interpolation to guide 3D representations (~\eg, radiance, SDF).
To generalize through scenes and to enable sparse-view inference, a limited number of source images (\eg,~$N=3$) are utilized to synthesize a target image during training. 
Along a ray emitted from a given pixel coordinate in the target image, $M$ 3D points are sampled as \( \{\mathbf{p}_m = {r}(t_m)=\textbf{o} + t_m\textbf{v},~m=1,...,M\} \), where \textbf{o} and \textbf{v} denote the ray origin and direction respectively.
To estimate the color and SDF value of 3D point $\mathbf{p}_m$, we acquire the corresponding image features and volume features.
We project $\mathbf{p}_m$ onto the source image $I_i$ as $\pi_i(\mathbf{p}_m)$ and obtain the corresponding image feature $\textbf{f}_i^\text{img}$ with bilinear interpolation.
In addition, the volume features $\textbf{f}^\text{vol}$ are obtained with trilinear interpolation at the normalized position. 
The color of the 3D position $\mathbf{c}_m$ is determined as a weighted sum of image pixel colors $I_i\left(\mathbf{p}_m\right) $, where the weights are estimated by $W(\cdot)$.
\begin{equation}
\label{eq:projection}
    \mathbf{c}_m = \sum_{i=1}^{N} W\left( \mathbf{f}^\text{vol}\left(\mathbf{p}_m\right), \{\mathbf{f}^\text{img}_i\left(\pi_i\left(\mathbf{p}_m\right)\right)\}^{N}_{i=1} \right)I_i\left(\mathbf{p}_m\right).
\end{equation}
The predicted 3D representations are aggregated into the 2D domain with volume rendering as follows:
\begin{equation}
\label{eq:nerf}
\begin{split}
    \hat{C}({r}) = & \sum_{m=1}^{M} T_m \alpha_m \mathbf{c}_m, \\
\end{split}
\end{equation}
where \( T_m = \prod_{k=1}^{m-1}(1-\alpha_k) \) denotes the accumulative transmittance, and \( \alpha_m \) indicates discrete opacity values defined as follows:
\begin{equation}
    \alpha_m = 1 - \exp\left(-\int_{t_m}^{t_{m+1}} \rho(t) dt\right),
\end{equation}
where $t_m \leq t_{m+1}$. The derivation of \( \rho(t) \) commonly follows the original definition in NeuS~\cite{neus} or VolSDF~\cite{volsdf}.


\subsection{Cross-View Matching Transformer} \label{cross-view transformer}
When input views become sparse, where the 3D space is under-constrained, it is important to identify correlations between the input views~\cite{sparf}.
Furthermore, as we assume that a view combination can be different, uncovering the relationships between input images and using them as reconstruction priors becomes more crucial for view-combination generalizability. 
While most existing neural surface reconstruction methods extract image features independently, recent works~\cite{unifyingflow, transmvsnet} have shown significant results by explicitly extracting cross-view correlation features of an input pair of images using transformer-based feature matching.

We initially extract multi-scale features $\boldsymbol{\phi}_i^l$ independently for each image $I_i$ using a Feature Pyramid Network (FPN)~\cite{fpn}. For each level $l$, the hierarchical features are then processed through a cross-view matching transformer $\mathcal{T}_M$ in a pairwise manner $F^l_i = \sum_{j\neq i}{\mathcal{T}_M(\boldsymbol{\phi}^l_i, \boldsymbol{\phi}^l_j)}$ for two purposes: 1) to construct correlation frustums that learn global implicit correlations among multi-view source images (Sec.~\ref{cross-view volume}), and 2) to extract explicit feature similarity scores (Sec.~\ref{similarity encoding}) for all source image pairs. Since both branches learn the correlation between source images, the framework seamlessly integrates the two parts with interdependent objectives, effectively serving as a global matching encoder. 

\subsection{Cascaded Cross-View Correlation Frustum} \label{cross-view volume}
Motivated by recent MVS methods that have shown the effectiveness of cost volume integration in generalizing sparse-view inference from arbitrary unseen scenes, we build a cascaded global volume with correlation features~\cite{transmvsnet} to provide cross-view information.
Our model constructs perspective frustums instead of regular Euclidean volumes for each source view by setting each view as a reference following C2F2NeUS~\cite{c2f2neus}.
For each reference image $I_i$, we align\Skip{ $N-1$} rest of the source images using differentiable warping to obtain total $N$ pairwise correlation frustums \( C^l_i \in \mathbb{R}^{1 \times d \times h \times w} \) for all levels $L$ as follows:
\vspace{-0.3cm}
\begin{equation}\label{eq:correlation}
    C^l_i = \sum_{\substack{j=1 \\ j \neq i}}^{N} \langle F^l_i, \hat{F}^l_{j \rightarrow i} \rangle, \quad l=1, \ldots, L
    \vspace{-0.3cm}
\end{equation}
where $F^l_i$ denotes reference-view feature and \( \hat{F}^l_{j \rightarrow i} \) denotes cross-view feature $F^l_j$ warped to $i$-th view from our cross-view matching transformer, and $d$, $h$, and $w$ denote the resolution of the volume.
We deliver the detailed aggregation of correlation volumes in the Appendix.
Multi-level correlation frustums are processed with 3D CNNs and output intermediate volumes $V_i^l \in \mathbb{R}^{ c \times d \times h\times w}$ and depths $D^l_i \in \mathbb{R}^{1 \times h \times w} $.
The intermediate volumes $V_i^l$, initially tailored for depth estimation, are transformed with additional 3D CNNs to generate ${V^\prime}_i^l$.
We output our global feature volumes by adding ${V^\prime}_i^l$ from all source views as ${V^\prime}^l=\sum_{i=1}^{N}{{V^\prime}^l_i}$, serving as a global correlation.
Our correlation frustum is different from C2F2NeUS~\cite{c2f2neus} in that we additionally utilize cross-view matching features from transformers and single-channel correlation frustums with much lower memory cost. Furthermore, we believe that correlation volumes guide our model to leverage the inherent geometrical relationships between different views extensively.

\subsection{Geometry-aware Similarity Encoding} \label{similarity encoding}
Recently, correspondence matching information on 2D has shown to be useful for 3D understanding~\cite{gmflow, matchnerf}.
To further guide our network for view-combination generalizability, we explicitly encode cosine similarity for all possible source pairs.
We use the extracted cross-view features for all $\binom{N}{2}$ pairs from the cross-view matching transformer. We only use the coarsest feature for similarity encoding.

To obtain the 2D cross-view features for the sampled 3D points $\mathbf{p}_m$, we project the 3D points onto the 2D cross-view matching features $F^0_{i}$ and $F^0_{j}$ and acquire the corresponding features $\textbf{f}_i=F^0_{i}\left(\pi_i\left(\mathbf{p}_m\right)\right)$ and $\textbf{f}_j=F^0_{j}\left(\pi_j\left(\mathbf{p}_m\right)\right)$.
After that, we calculate the cosine similarity of the two vectors  $\hat{\textbf{s}} = \cos(\textbf{f}_i, \textbf{f}_j)$ in a group-wise manner \cite{gwcnet}.
We average the cosine similarities for all pairs and obtain encoded similarity vector $\textbf{f}_s$, allowing the dimension invariant to input source images. 
Given an arbitrary view combination for inference, the explicit similarity scores provide a robust geometry prior, regularizing the feature space for matching.




\subsection{Reconstruction Transformers} \label{aggregation transformer}

\textbf{View-aggregation Transformer.} 
We combine global correlation from our correlation frustums, cross-view matching features, and similarity scores using our view-aggregation transformer $\mathcal{T}_\text{A}$.
We acquire 2D cross-view matching features and similarity scores for all sampled points in the 3D space by projecting the points onto each image view and applying bilinear interpolation. 
Additionally, we employ trilinear interpolation on the points to feature volumes ${V^\prime}^l$ to obtain global correlation features.
Finally, we add a learnable token feature $\textbf{f}_0$ to capture global consistent rendering features as in ~\cite{retr, volrecon}.
All features above are aggregated via linear self-attention transformers~\cite{transformers, attention}:
\begin{equation}\label{viewtrans}
    \textbf{f}^p, \{\textbf{f}^{\prime}_i \}_{i=1}^{N} = \mathcal{T}_\text{A}(\textbf{f}_0, \{\textbf{f}_i\}_{i=1}^{N}, \textbf{f}_v, \textbf{f}_s),
\end{equation}
where $\textbf{f}^p$ is a projection vector and $\{\textbf{f}^{\prime}_i \}_{i=1}^{N}$ are aggregated features further used for rendering and $\textbf{f}_v$ denotes the concatenated volume features sampled from ${V^\prime}^l$ for all levels.

\noindent
\textbf{Geometry-aware Ray Transformer.}
To estimate implicit geometry representation (~\eg,SDF) along the rays, we first apply positional encoding to the points along the ray. 
This is achieved by calculating the difference between the camera coordinates of sampled points and the intermediate depth obtained from the cascaded correlation frustum as in \cite{mvsnet}.
We then apply positional encoding~\cite{transformers} $\gamma(\cdot)$ to embed the order of the ray and to extract geometry-aware features $\textbf{f}^{\prime}_i$ with the self-attention linear transformer $\mathcal{T}_\text{R}$ as follows:
\begin{equation}\label{raytrans}
    \{\textbf{f}^{\prime}_i \}_{i=1}^{M} = \mathcal{T}_\text{R}(\textbf{f}_p, \text{cat}(\gamma(z_i-z_d))^{M}_{i=1}).
\end{equation}
where $z_i$ denotes depth values of a $i$-th sampled point in the camera coordinate and $z_d$ denotes the intermediate depth value of the ray.  
We utilize color blending to obtain color value for each point following \cite{volrecon} and aggregate all points along the ray with volume rendering as in Eq.~\ref{eq:nerf}. We adopt the method of NeuS \cite{neus} to volume render SDF.

\subsection{Training of UFORecon} \label{Training}
\textbf{Random Set Training.}
First, we follow the common protocol of training with a pre-determined best view combination based on the view-selection score~\cite{mvsnet} and expect generalization to arbitrary or unfavorable camera configurations not seen during training.
Beyond the existing training strategy, we suggest a random set training, which randomly selects $N$ source views in the training phase, expecting the network to handle arbitrary view combinations. A detailed explanation and evidence are presented in the Appendix.

\noindent
\textbf{Loss function.}
Our loss function is defined as a weighted sum of two loss functions as:
\begin{equation}\label{loss}
    \mathcal{L} = \mathcal{L}_{color} + \alpha \mathcal{L}_{depth},
\end{equation}
where $\mathcal{L}_{color}$ is formulated to minimize the ground truth colors and rendered color.
Similarly, $\mathcal{L}_{depth}$ is formulated to minimize the ground truth depths and rendered depth, following VolRecon~\cite{volrecon}.

\section{Experiments}

\begin{table*}[t!]
\centering
\small
\resizebox{0.9\textwidth}{!}{
\begin{tabular}{lccccccccccccccccc}
\hline
Scan & 24 & 37 & 40 & 55 & 63 & 65 & 69 & 83 & 97 & 105 & 106 & 110 & 114 & 118 & 122 & Mean (CD) $\downarrow$ \\
\hline \hline
COLMAP~\cite{colmap} & 0.90 & 2.89 & 1.63 & 1.08 & 2.18 & 1.94 & 1.61 & 1.30 & 2.34 & 1.28 & 1.10 & 1.42 & 0.76 & 1.17 & 1.14 & 1.52 \\

TransMVSNet~\cite{transmvsnet} & 1.07 & 3.14 & 2.39 & 1.30 & 1.35 & 1.61 & 0.73 & 1.60 & 1.15 & 0.94 & 1.34 & 0.46 & 0.60 & 1.20 & 1.46 & 1.35 \\
\hline
VolSDF~\cite{volsdf} & 4.03 & 4.21 & 6.12 & 1.63 & 3.24 & 2.73 & 2.84 & 1.63 & 5.14 & 3.09 & 2.08 & 4.81 & 0.60 & 3.51 & 2.18 & 3.41 \\
NeuS~\cite{neus} & 4.57 & 4.49 & 3.97 & 4.32 & 4.63 & 1.95 & 4.64 & 3.83 & 5.40 & 5.60 & 6.47 & 6.68 & 2.96 & 5.57 & 6.11 & 4.90 \\
SparseNeuS-ft~\cite{sparseneus} & 1.29 & 2.27 & 1.57 & 0.88 & 1.61 & 1.86 & 1.06 & 1.27 & 1.42 & 1.07 & 0.99 & 0.87 & 0.54 & 1.15 & 1.18 & 1.27 \\
\hline
PixelNeRF~\cite{pixelnerf} & 5.13 & 8.07 & 5.85 & 4.40 & 7.11 & 4.64 & 5.68 & 6.76 & 9.05 & 6.11 & 3.95 & 5.92 & 6.26 & 6.89 & 6.93 & 6.28 \\
IBRNet~\cite{ibrnet} & 2.29 & 3.70 & 2.66 & 1.83 & 3.02 & 2.83 & 1.77 & 2.28 & 2.73 & 1.96 & 1.87 & 2.13 & 1.58 & 2.05 & 2.09 &  2.32 \\
MVSNeRF~\cite{mvsnerf} & 1.96 & 3.27 & 2.54 & 1.93 & 2.57 & 2.71 & 1.82 & 1.72 & 2.29 & 1.75 & 1.72 & 1.47 & 1.29 & 2.09 & 2.26 &  2.09 \\
\hline
SparseNeuS~\cite{sparseneus} & 1.68 & 3.06 & 2.25 & 1.10 & 2.37 & 2.18 & 1.28 & 1.47 & 1.80 & 1.23 & 1.19 & 1.17 & 0.75 & 1.56 & 1.55 &  1.64 \\
VolRecon~\cite{volrecon} & 1.20 & 2.59 & 1.56 & 1.08 & 1.43 & 1.92 & 1.11 & 1.48 & 1.42 & 1.05 & 1.19 & 1.38 & 0.74 & 1.23 & 1.27 &  1.38 \\
ReTR~\cite{retr} & 1.05 & 2.31 & 1.44 & 0.98 & 1.18 & 1.52 & 0.88 & 1.35 & 1.30 & 0.87 & 1.07 & 0.77 & 0.59 & 1.05 & 1.12 &  1.17 \\
C2F2NeuS~\cite{c2f2neus} & 1.12 & 2.42 & 1.40 & \textbf{0.75} & 1.41 & 1.77 & 0.85 & \textbf{1.16} & 1.26 & \underline{0.76} & 0.91 & 0.60 & \textbf{0.46} & 0.88 & \textbf{0.92} &  1.11 \\
Ours & \textbf{0.76} & \textbf{2.05} & \textbf{1.31} & \underline{0.82} & \textbf{1.12} & \textbf{1.18} & \underline{0.74} & \underline{1.17} & \textbf{1.11} & \textbf{0.71} & \textbf{0.88} & \underline{0.58} & 0.54 & \textbf{0.86} & 0.99 &  \textbf{0.99} \\
Ours* & \underline{0.77} & \underline{2.10} & \underline{1.34} & 0.87 & \underline{1.15} & \underline{1.16} & \textbf{0.71} & 1.25 & \underline{1.17} & 0.81 & \underline{0.90} & \textbf{0.57} & \underline{0.51} & \textbf{0.86} & \underline{0.97} & \underline{1.01} \\
\hline
\end{tabular}
}
\caption{\textbf{A quantitative results on favorable sets.} Our method outperforms traditional 3D reconstruction methods (first section), per-scene surface reconstruction methods (second section), generalizable novel-view synthesis methods (third section), and generalizable surface reconstruction methods (fourth section). The \textbf{Bold} numbers mean the best, and the \underline{underlined} values indicate the second-best scores. (*) denotes the use of a random set training strategy.}
\label{table:fav_comparison}
\end{table*}

\begin{table*}[t!]
\centering
\small
\resizebox{0.85\textwidth}{!}{
\begin{tabular}{lccccccccccccccccc}
\hline
Scan & 24 & 37 & 40 & 55 & 63 & 65 & 69 & 83 & 97 & 105 & 106 & 110 & 114 & 118 & 122 & Mean (CD) $\downarrow$ \\
\hline \hline
TransMVSNet~\cite{transmvsnet} & 8.32 & 8.68 & 7.46 & 8.16 & 8.22 & 6.74 & 7.74 & 8.36 & 10.67 & 9.70 & 8.86 & 7.99 & 8.32 & 8.90 & 7.13 & 8.35 \\
\hline
SparseNeuS~\cite{sparseneus} & 5.24 & 5.00 & 5.76 & 4.95 & 3.80 & 4.12 & 3.87 & 3.44 & 3.48 & 3.23 & 4.61 & 4.24 & 2.13 & 4.04 & 4.46 & 4.16 \\
VolRecon~\cite{volrecon} & 3.43 & 3.64 & 4.26 & 4.63 & 2.43 & 3.40 & 2.81 & 2.41 & 2.36 & 2.49 & 3.79 & 3.55 & 1.44 & 3.60 & 3.38 &  3.18 \\
ReTR~\cite{retr} & 3.00 & 3.98 & 3.78 & 4.22 & 2.22 & 2.93 & 3.00 & 2.51 & 2.24 & 2.36 & 2.36 & 3.92 & 1.63 & 2.83 & 3.07 &  2.94 \\
Ours & \underline{1.39} & \underline{2.25} & \underline{1.65} & \underline{1.96} & \underline{1.53} & \underline{1.61} & \underline{1.22} & \underline{1.92} & \underline{1.36} & \underline{1.66} & \underline{1.75} & \underline{1.29} & \underline{0.73} & \underline{1.70} & \underline{1.39} &  \underline{1.56} \\
Ours* & \textbf{1.31} & \textbf{2.00} & \textbf{1.41} & \textbf{1.36} & \textbf{1.24} & \textbf{1.58} & \textbf{1.06} & \textbf{1.44} & \textbf{1.37} & \textbf{0.99} & \textbf{1.45} & \textbf{0.96} & \textbf{0.58} & \textbf{1.34} & \textbf{1.09} &  \textbf{1.28} \\
\hline
\end{tabular}
}
\caption{\textbf{A quantitative results on unfavorable sets.} (*) denotes
the use of a random set training strategy.}
\label{table:unfav_comparison}
\end{table*}

In this section, we first explain our experimental settings, including datasets, baseline, and implementation details (Sec.~\ref{ssec:5.1}).
Second, we qualitatively and quantitatively evaluate our method on generalizable surface reconstruction mainly focusing on view-combination generalizability (Sec.~\ref{ssec:results}).
Last, we analyze the effectiveness of our individual elements (Sec.~\ref{ssec:analysis}).

\subsection{Experimental Settings}
\label{ssec:5.1}
\vspace{0.5ex}\noindent
\textbf{Datasets.}
We report our results on the DTU datasets~\cite{dtu}, for the challenging scenario of 3 input views in unseen scenes following \cite{sparseneus,volrecon,retr,c2f2neus}.
The DTU dataset comprises scenes captured from 49 distinct frontal viewpoints, each accompanied by its respective camera matrix.
We use 3 source views from each test scene to build a combination and evaluate the generalization ability of our method.  
For all possible view combinations, we calculate the VC scores and rank them based on the scores in descending order. Subsequently, we evenly divide these combinations into three groups: \textit{favorable}, \textit{normal}, and \textit{unfavorable}.

\label{sssec:baseline}
\vspace{0.5ex}\noindent
\textbf{Baseline.}
To demonstrate the effectiveness of our method, we mainly compare it with (1) SparseNeus~\cite{sparseneus}, VolRecon ~\cite{volrecon}, ReTR~\cite{retr}, and C2F2NeuS~\cite{c2f2neus} the state-of-the-art generalizable neural implicit reconstruction method. 
For evaluating scene and viewpoints generalizability, we additionally compare with (2) Generalizable neural rendering methods \cite{pixelnerf,mvsnerf,ibrnet}, and (3) Neural implicit reconstruction \cite{volsdf,neus} which require per-scene training. Additionally, we include (4) traditional MVS methods~\cite{colmap,transmvsnet}.

\begin{figure*}[t]\label{qualitative}
    \centering
    \includegraphics[width=0.9\linewidth, trim={0cm 11.5cm 13.8cm 0cm}, clip]{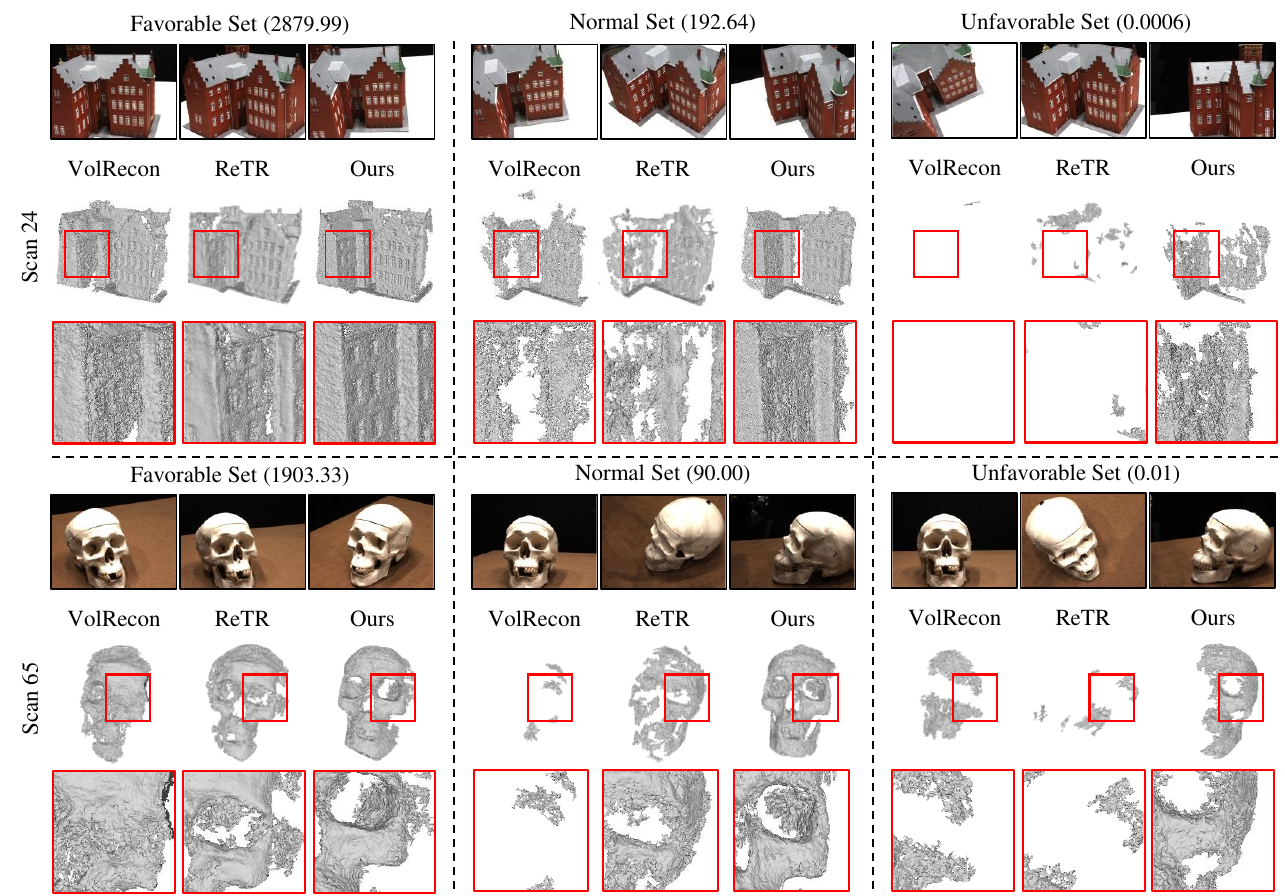}
    \caption{\textbf{A qualitative results of Surface Reconstruction across various VC Levels.} The numbers in parentheses denote the view-combination score. Our method consistently outperforms previous methods at all levels and in different scenes. More qualitative results can be found in the Appendix.}
    \label{fig:qualitative}
\end{figure*}

\begin{figure}
    \centering 
    \includegraphics[width=\linewidth, trim={0 7.3cm 2.6cm 0.0cm}, clip]{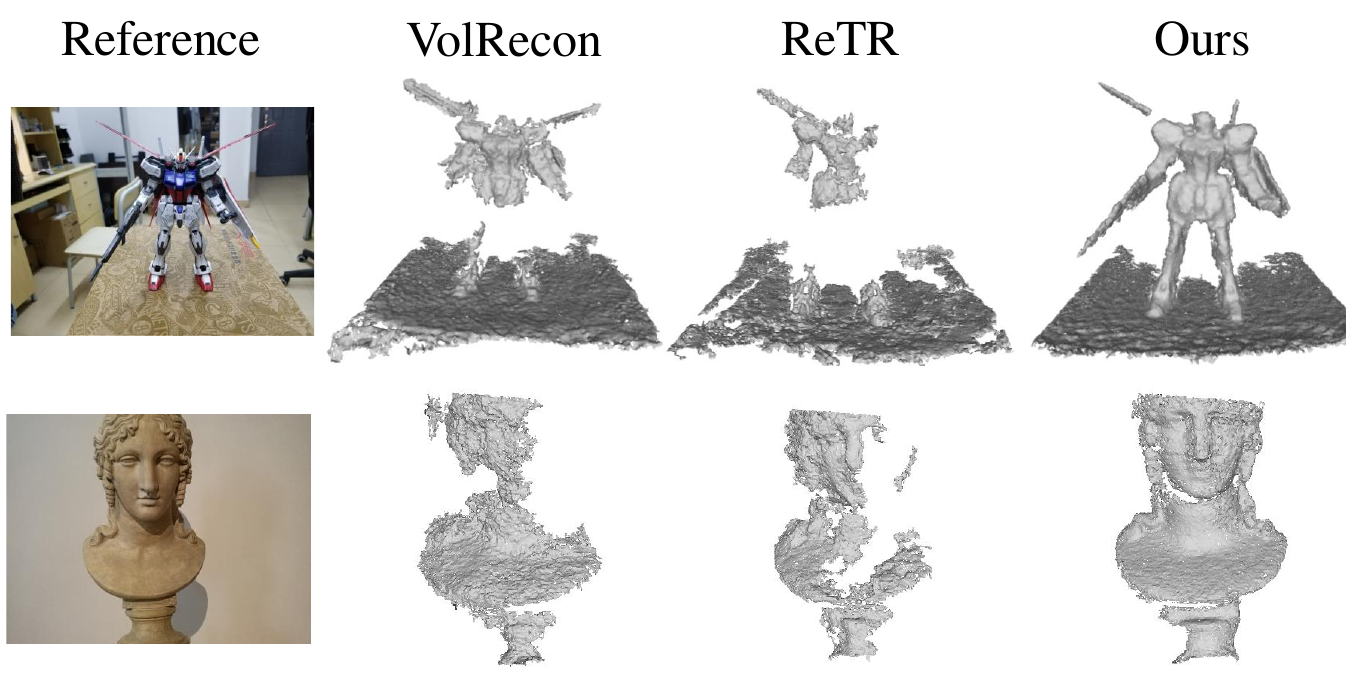} 
    \caption{\textbf{Unfavorable 4-views tests on BlendedMVS dataset.}}
    \vspace{-0.3cm}
\label{fig:qualitative_blendedmvs}
\end{figure}

\begin{figure}[t]
    \centering
    \includegraphics[width=0.8\linewidth, trim={0cm 14.5cm 13cm 0cm}, clip]{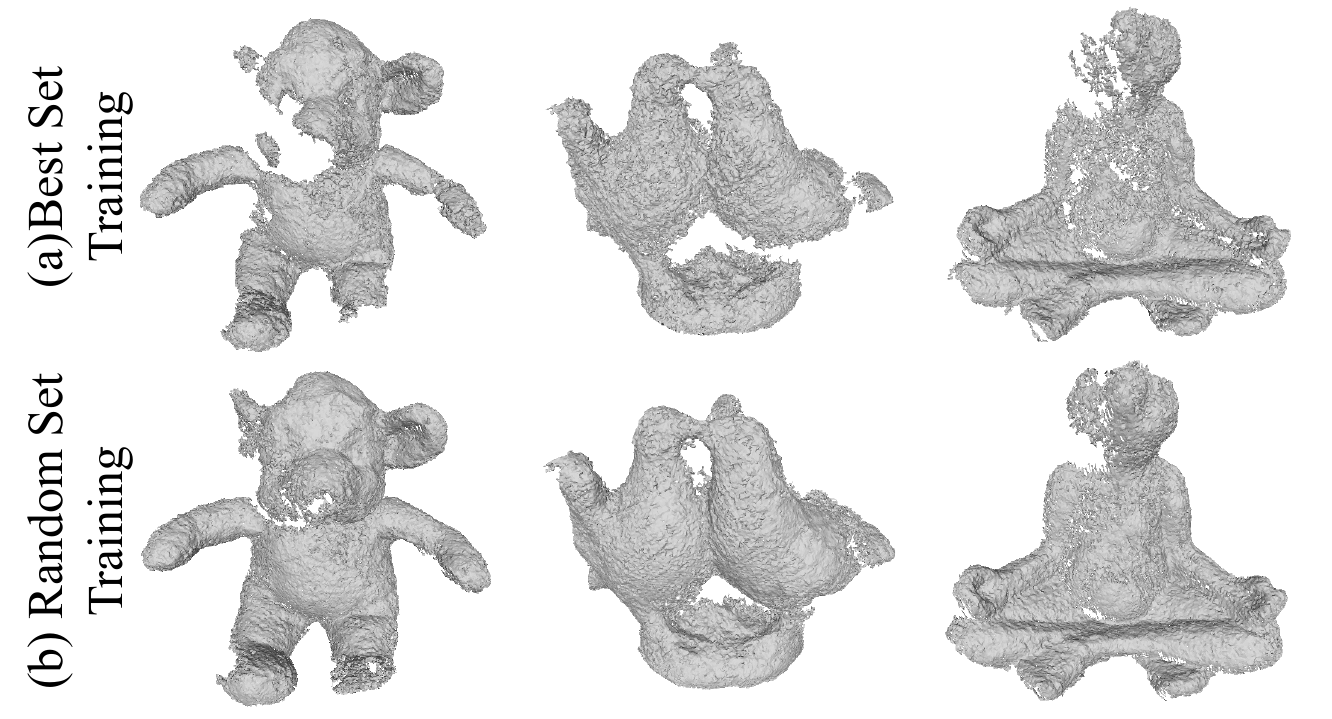}
    \caption{\textbf{Effect of random set training on unfavorable sets.}}
    \vspace{-0.4cm}
\label{fig:best_random}
\end{figure}

\label{sssec:implementation}
\vspace{0.5ex}\noindent
\textbf{Implementation Details.}
In training, we take $N=4$ source images and one target image with the resolution of $640 \times 512$.
\Skip{In sampling the explicit feature similarity, we set the number of groups to 4, and we only use the coarsest level of features.}
We train the network in an end-to-end manner with Adam optimizer \cite{adam} on a single RTX 4090 GPU. 
The learning rate is set to 1e-4. We use a hierarchical sampling strategy in both training and testing with 64 points for both coarse and fine sampling. For global feature volume, we use $L=3$ for cascaded strategy and follow the resolution of TransMVSNet~\cite{transmvsnet} and set the number of depth hypotheses for each level to 48, 32, and 8 respectively. More implementation details can be found in the Appendix.

\subsection{Evaluation on Surface Reconstruction}
\label{ssec:results}
Chamfer Distance (CD) is used as an evaluation metric in all experiments. 
Note that for fair comparisons, our training adheres to the pre-determined best view combinations, aligning with the existing protocols~\cite{sparseneus,volrecon,retr,c2f2neus} unless otherwise specified. 

\noindent
\textbf{Reconstruction on Favorable Sets.} 
Following the existing best-selected test protocol~\cite{sparseneus,volrecon,retr,c2f2neus}, we evaluate with the fixed favorable views for all test scenes.
As shown in Table~\ref{table:fav_comparison}, our approach outperforms the state-of-the-art models even in the conventional setting.
It implies that the superiority of our model is also consistent in all \textit{VC} score levels.

\vspace{0.5ex}\noindent
\textbf{Reconstruction on Unfavorable Sets.}
To analyze the proposed view-combination generalizability, we test with \textit{unfavorable} sets.
To this end, we sample a view set consisting of 3 images from the unfavorable view group and use the same set for all 15 test scenes. 
As shown in Table \ref{table:unfav_comparison}, our method consistently achieves significantly better performance in all scenes.
Our method achieves better view-combination generalizability by capturing robust prior for image combination.
In addition, we qualitatively compare our method in various \textit{VC} score levels.
As shown in Figure \ref{fig:qualitative}, ours shows much better reconstruction quality compared to existing works, particularly showing huge advantages in unfavorable sets. 
We further tested with unfavorable sets of BlendedMVS dataset without fine-tuning as shown in Figure \ref{fig:qualitative_blendedmvs}. It shows that our method is generalizable to unfavorable view-combinations of other datasets.

\subsection{Analysis on UFORecon}
\label{ssec:analysis}

\noindent
\textbf{Ablation Study.}
We first perform ablation experiments to assess the impact of each component within our approach.
Table \ref{tab:ablation_study} shows the performance in an unfavorable scenario as specific components are removed from the framework.
Each component contributes to the overall performance. Each level of correlation frustum contributes to better performance.
Encoding similarity further improves the performance even with a full level of correlation frustums.
Also, note that depth supervision plays an important role in our framework similar to \cite{volrecon,retr}. 

\noindent
\textbf{Analysis on Correlation Frustums.}
A higher level of frustum encodes relatively finer and local details while a lower level encodes global geometry features.
As shown in Table~\ref{tab:ablation_study}, the performance degradation is more severe when excluding lower levels.


\begin{table}[t!]
\centering
\small
\resizebox{0.55\linewidth}{!}{
\begin{tabular}{c|c}
\hline
\textbf{Methods} & Unfavorable \\ \hline
w/o CF & 3.19\\
CF ($L=1$) & 1.91 \\
CF ($L=2$) & 1.71 \\
w/o similarity encoding & 1.62 \\
w/o $L_{depth}$ & 2.26 \\
UFORecon & 1.56 \\ \hline
\end{tabular}
}
\caption{\textbf{Ablation study on each element in UFORecon.} CF denotes cross-view correlation frustum. All experiments are tested with unfavorable sets.}
\label{tab:ablation_study}
\vspace{-0.2cm}
\end{table}

\begin{table}[t!]
\centering
\small
\resizebox{\linewidth}{!}{
\begin{tabular}{c|c c|c}
\hline
Method & Favorable & Unfavorable & \#params \\
\hline \hline
VolRecon~\cite{volrecon} & 1.38 & 3.18 & 0.88 M \\ 
VolRecon~\cite{volrecon} + MatchNeRF~\cite{matchnerf} & 1.26 & 2.47 & 4.97 M \\ 
Ours & 0.99 & 1.56 & 1.5 M \\ 
\hline
\end{tabular}
}
\caption{\textbf{Effect of explicit similarity score.} We combine two representative baselines in generalizable surface reconstruction~\cite{volrecon} and correspondence matching-based NVS method~\cite{matchnerf}. This hybrid model shows 10\% improvements in favorable sets and 22\% in unfavorable sets.
Our method outperforms this hybrid model with better efficiency.}
\label{tab:ablation_matchnerf}
\end{table}

\noindent
\textbf{Analysis on Explicit Similarity Encoding.}
We conduct a focused analysis on the impact of similarity encoding by contrasting our approach with a hybrid model. 
Specifically, we integrate MatchNeRF~\cite{matchnerf}, known for its utilization of explicit similarity scores for generalizable rendering, with VolRecon~\cite{volrecon}.
This combined model is then tested on both favorable and unfavorable datasets.
As indicated in Table \ref{tab:ablation_matchnerf}, incorporating explicit feature similarity notably enhances the quality of generalizable surface reconstructions especially in unfavorable conditions.
This reflects that explicitly feature similarity provides robust view-combination prior.
Nevertheless, our proposed approach outperforms in both favorable and unfavorable conditions, effectively capturing correlation features with much fewer parameters.

\noindent  
\textbf{Analysis on Random Set Training.}
We analyze the effect of the random set training. We expect the network to generalize better to various combinations without losing accuracy. As shown in Figure~\ref{fig:cd_fav}, it consistently improves the reconstruction quality at all \textit{VC} score levels. 
In unfavorable sets, random set training shows 17\% of improvements as shown in Table~\ref{table:unfav_comparison}.
Qualitatively, random set training consistently achieves more complete reconstructions as shown in Figure~\ref{fig:best_random}.
However, we find that directly applying random set training on the previous method~\cite{volrecon} destabilizes training dynamics and leads to a suboptimal result.
We conjecture that utilizing cross-view correlation features guides the network to stably improve the reconstruction performance.

\section{Conclusion}
In this paper, we have introduced a novel concept of \textit{view-combination generalizability} in the field of generalizable implicit surface reconstruction.
Our analysis showed that existing methods tend to overfit to specific learned view combinations, resulting in suboptimal performance with unfavorable combinations.
To address this challenge, we have proposed a novel framework, UFORecon, that leverages cross-view matching features to learn correlation among input images.
This is achieved by building correlation frustums with cross-view features and encoding explicit 2D matching similarities across all source image pairs.
Notably, our method demonstrates superior performance over previous surface reconstruction methods, not only in unfavorable view combinations but also in favorable ones.

\noindent
\textbf{Limitations and future work.}
Although our method achieves advancements in view-combination generalizability for surface reconstruction, it is still challenging to reconstruct scenes with much higher complexity.
In addition, we assume that the ground-truth camera poses are given,  
We include a detailed discussion on this in the Appendix. 
We hope that our contributions will inspire further research to extend this to such directions.

\noindent
\textbf{Acknowledgement}.
This work was supported by the National Research Foundation of Korea(NRF) grant funded by the Korea government(MSIT) (No. RS-2023-00208506(2024)).
Prof. Sung-Eui Yoon is a corresponding author.


{
    \small
    \bibliographystyle{ieeenat_fullname}
    \bibliography{main}

\begin{thebibliography}{54}
\providecommand{\natexlab}[1]{#1}
\providecommand{\url}[1]{\texttt{#1}}
\expandafter\ifx\csname urlstyle\endcsname\relax
  \providecommand{\doi}[1]{doi: #1}\else
  \providecommand{\doi}{doi: \begingroup \urlstyle{rm}\Url}\fi

\bibitem[Aan{\ae}s et~al.(2016)Aan{\ae}s, Jensen, Vogiatzis, Tola, and Dahl]{dtu}
Henrik Aan{\ae}s, Rasmus~Ramsb{\o}l Jensen, George Vogiatzis, Engin Tola, and Anders~Bjorholm Dahl.
\newblock Large-scale data for multiple-view stereopsis.
\newblock \emph{International Journal of Computer Vision}, 120:\penalty0 153--168, 2016.

\bibitem[Campbell et~al.(2008)Campbell, Vogiatzis, Hern{\'a}ndez, and Cipolla]{depthhypotheses}
Neill~DF Campbell, George Vogiatzis, Carlos Hern{\'a}ndez, and Roberto Cipolla.
\newblock Using multiple hypotheses to improve depth-maps for multi-view stereo.
\newblock In \emph{Computer Vision--ECCV 2008: 10th European Conference on Computer Vision, Marseille, France, October 12-18, 2008, Proceedings, Part I 10}, pages 766--779. Springer, 2008.

\bibitem[Chang et~al.(2022)Chang, Bo{\v{z}}i{\v{c}}, Zhang, Yan, Chen, S{\"u}sstrunk, and Nie{\ss}ner]{rc-mvsnset}
Di Chang, Alja{\v{z}} Bo{\v{z}}i{\v{c}}, Tong Zhang, Qingsong Yan, Yingcong Chen, Sabine S{\"u}sstrunk, and Matthias Nie{\ss}ner.
\newblock Rc-mvsnet: unsupervised multi-view stereo with neural rendering.
\newblock In \emph{European Conference on Computer Vision}, pages 665--680. Springer, 2022.

\bibitem[Chen et~al.(2021)Chen, Xu, Zhao, Zhang, Xiang, Yu, and Su]{mvsnerf}
Anpei Chen, Zexiang Xu, Fuqiang Zhao, Xiaoshuai Zhang, Fanbo Xiang, Jingyi Yu, and Hao Su.
\newblock Mvsnerf: Fast generalizable radiance field reconstruction from multi-view stereo.
\newblock In \emph{Proceedings of the IEEE/CVF International Conference on Computer Vision}, pages 14124--14133, 2021.

\bibitem[Chen et~al.(2023)Chen, Xu, Wu, Zheng, Cham, and Cai]{matchnerf}
Yuedong Chen, Haofei Xu, Qianyi Wu, Chuanxia Zheng, Tat-Jen Cham, and Jianfei Cai.
\newblock Explicit correspondence matching for generalizable neural radiance fields.
\newblock \emph{arXiv preprint arXiv:2304.12294}, 2023.

\bibitem[Cheng et~al.(2023)Cheng, Cao, and Shan]{idpose}
Weihao Cheng, Yan-Pei Cao, and Ying Shan.
\newblock Id-pose: Sparse-view camera pose estimation by inverting diffusion models.
\newblock \emph{arXiv preprint arXiv:2306.17140}, 2023.

\bibitem[Dai et~al.(2023)Dai, Zhu, Geng, Ruan, Zhang, and Wang]{graspnerf}
Qiyu Dai, Yan Zhu, Yiran Geng, Ciyu Ruan, Jiazhao Zhang, and He Wang.
\newblock Graspnerf: Multiview-based 6-dof grasp detection for transparent and specular objects using generalizable nerf.
\newblock In \emph{2023 IEEE International Conference on Robotics and Automation (ICRA)}, pages 1757--1763. IEEE, 2023.

\bibitem[Deitke et~al.(2023)Deitke, Schwenk, Salvador, Weihs, Michel, VanderBilt, Schmidt, Ehsani, Kembhavi, and Farhadi]{deitke2023objaverse}
Matt Deitke, Dustin Schwenk, Jordi Salvador, Luca Weihs, Oscar Michel, Eli VanderBilt, Ludwig Schmidt, Kiana Ehsani, Aniruddha Kembhavi, and Ali Farhadi.
\newblock Objaverse: A universe of annotated 3d objects.
\newblock In \emph{Proceedings of the IEEE/CVF Conference on Computer Vision and Pattern Recognition}, pages 13142--13153, 2023.

\bibitem[Ding et~al.(2022)Ding, Yuan, Zhu, Zhang, Liu, Wang, and Liu]{transmvsnet}
Yikang Ding, Wentao Yuan, Qingtian Zhu, Haotian Zhang, Xiangyue Liu, Yuanjiang Wang, and Xiao Liu.
\newblock Transmvsnet: Global context-aware multi-view stereo network with transformers.
\newblock In \emph{Proceedings of the IEEE/CVF Conference on Computer Vision and Pattern Recognition}, pages 8585--8594, 2022.

\bibitem[Fan et~al.(2023)Fan, Pan, Wang, Jiang, Jiang, Xu, Zhu, Wang, and Wang]{posefreegen}
Zhiwen Fan, Panwang Pan, Peihao Wang, Yifan Jiang, Hanwen Jiang, Dejia Xu, Zehao Zhu, Dilin Wang, and Zhangyang Wang.
\newblock Pose-free generalizable rendering transformer.
\newblock \emph{arXiv e-prints}, pages arXiv--2310, 2023.

\bibitem[Furukawa and Ponce(2009)]{stereopsis}
Yasutaka Furukawa and Jean Ponce.
\newblock Accurate, dense, and robust multiview stereopsis.
\newblock \emph{IEEE transactions on pattern analysis and machine intelligence}, 32\penalty0 (8):\penalty0 1362--1376, 2009.

\bibitem[Galliani et~al.(2015)Galliani, Lasinger, and Schindler]{parallelmvs}
Silvano Galliani, Katrin Lasinger, and Konrad Schindler.
\newblock Massively parallel multiview stereopsis by surface normal diffusion.
\newblock In \emph{Proceedings of the IEEE International Conference on Computer Vision}, pages 873--881, 2015.

\bibitem[Gu et~al.(2020)Gu, Fan, Zhu, Dai, Tan, and Tan]{casmvsnet}
Xiaodong Gu, Zhiwen Fan, Siyu Zhu, Zuozhuo Dai, Feitong Tan, and Ping Tan.
\newblock Cascade cost volume for high-resolution multi-view stereo and stereo matching.
\newblock In \emph{Proceedings of the IEEE/CVF conference on computer vision and pattern recognition}, pages 2495--2504, 2020.

\bibitem[Guo et~al.(2019)Guo, Yang, Yang, Wang, and Li]{gwcnet}
Xiaoyang Guo, Kai Yang, Wukui Yang, Xiaogang Wang, and Hongsheng Li.
\newblock Group-wise correlation stereo network.
\newblock In \emph{Proceedings of the IEEE/CVF conference on computer vision and pattern recognition}, pages 3273--3282, 2019.

\bibitem[H{\"a}ne et~al.(2015)H{\"a}ne, Sattler, and Pollefeys]{driving}
Christian H{\"a}ne, Torsten Sattler, and Marc Pollefeys.
\newblock Obstacle detection for self-driving cars using only monocular cameras and wheel odometry.
\newblock In \emph{2015 IEEE/RSJ International Conference on Intelligent Robots and Systems (IROS)}, pages 5101--5108. IEEE, 2015.

\bibitem[Ji et~al.(2017)Ji, Gall, Zheng, Liu, and Fang]{surfacenet}
Mengqi Ji, Juergen Gall, Haitian Zheng, Yebin Liu, and Lu Fang.
\newblock Surfacenet: An end-to-end 3d neural network for multiview stereopsis.
\newblock In \emph{Proceedings of the IEEE international conference on computer vision}, pages 2307--2315, 2017.

\bibitem[Jiang et~al.(2023)Jiang, Jiang, Zhao, and Huang]{leap}
Hanwen Jiang, Zhenyu Jiang, Yue Zhao, and Qixing Huang.
\newblock Leap: Liberate sparse-view 3d modeling from camera poses.
\newblock \emph{arXiv preprint arXiv:2310.01410}, 2023.

\bibitem[Jiang et~al.(2020)Jiang, Ji, Han, and Zwicker]{sdfdiff}
Yue Jiang, Dantong Ji, Zhizhong Han, and Matthias Zwicker.
\newblock Sdfdiff: Differentiable rendering of signed distance fields for 3d shape optimization.
\newblock In \emph{Proceedings of the IEEE/CVF conference on computer vision and pattern recognition}, pages 1251--1261, 2020.

\bibitem[Kar et~al.(2017)Kar, H{\"a}ne, and Malik]{lsm}
Abhishek Kar, Christian H{\"a}ne, and Jitendra Malik.
\newblock Learning a multi-view stereo machine.
\newblock \emph{Advances in neural information processing systems}, 30, 2017.

\bibitem[Katharopoulos et~al.(2020)Katharopoulos, Vyas, Pappas, and Fleuret]{transformers}
Angelos Katharopoulos, Apoorv Vyas, Nikolaos Pappas, and Fran{\c{c}}ois Fleuret.
\newblock Transformers are rnns: Fast autoregressive transformers with linear attention.
\newblock In \emph{International conference on machine learning}, pages 5156--5165. PMLR, 2020.

\bibitem[Kingma and Ba(2015)]{adam}
Diederik Kingma and Jimmy Ba.
\newblock Adam: A method for stochastic optimization.
\newblock In \emph{International Conference on Learning Representations (ICLR)}, San Diega, CA, USA, 2015.

\bibitem[Kutulakos and Seitz(2000)]{carving}
Kiriakos~N Kutulakos and Steven~M Seitz.
\newblock A theory of shape by space carving.
\newblock \emph{International journal of computer vision}, 38:\penalty0 199--218, 2000.

\bibitem[Lhuillier and Quan(2005)]{quasi}
Maxime Lhuillier and Long Quan.
\newblock A quasi-dense approach to surface reconstruction from uncalibrated images.
\newblock \emph{IEEE transactions on pattern analysis and machine intelligence}, 27\penalty0 (3):\penalty0 418--433, 2005.

\bibitem[Li et~al.(2021)Li, Liu, Drenkow, Ding, Creighton, Taylor, and Unberath]{sttr}
Zhaoshuo Li, Xingtong Liu, Nathan Drenkow, Andy Ding, Francis~X Creighton, Russell~H Taylor, and Mathias Unberath.
\newblock Revisiting stereo depth estimation from a sequence-to-sequence perspective with transformers.
\newblock In \emph{Proceedings of the IEEE/CVF international conference on computer vision}, pages 6197--6206, 2021.

\bibitem[Li et~al.(2023)Li, Li, and Zhu]{read}
Zhuopeng Li, Lu Li, and Jianke Zhu.
\newblock Read: Large-scale neural scene rendering for autonomous driving.
\newblock In \emph{Proceedings of the AAAI Conference on Artificial Intelligence}, pages 1522--1529, 2023.

\bibitem[Liang et~al.(2023)Liang, He, and Chen]{retr}
Yixun Liang, Hao He, and Ying-cong Chen.
\newblock Rethinking rendering in generalizable neural surface reconstruction: A learning-based solution.
\newblock \emph{arXiv preprint arXiv:2305.18832}, 2023.

\bibitem[Lin et~al.(2017)Lin, Doll{\'a}r, Girshick, He, Hariharan, and Belongie]{fpn}
Tsung-Yi Lin, Piotr Doll{\'a}r, Ross Girshick, Kaiming He, Bharath Hariharan, and Serge Belongie.
\newblock Feature pyramid networks for object detection.
\newblock In \emph{Proceedings of the IEEE conference on computer vision and pattern recognition}, pages 2117--2125, 2017.

\bibitem[Long et~al.(2022)Long, Lin, Wang, Komura, and Wang]{sparseneus}
Xiaoxiao Long, Cheng Lin, Peng Wang, Taku Komura, and Wenping Wang.
\newblock Sparseneus: Fast generalizable neural surface reconstruction from sparse views.
\newblock In \emph{European Conference on Computer Vision}, pages 210--227. Springer, 2022.

\bibitem[Lorensen and Cline(1987)]{marchingcubes}
William~E Lorensen and Harvey~E Cline.
\newblock Marching cubes: A high resolution 3d surface construction algorithm.
\newblock \emph{ACM SIGGRAPH Computer Graphics}, 21\penalty0 (4):\penalty0 163--169, 1987.

\bibitem[Middelberg et~al.(2014)Middelberg, Sattler, Untzelmann, and Kobbelt]{scalable}
Sven Middelberg, Torsten Sattler, Ole Untzelmann, and Leif Kobbelt.
\newblock Scalable 6-dof localization on mobile devices.
\newblock In \emph{Computer Vision--ECCV 2014: 13th European Conference, Zurich, Switzerland, September 6-12, 2014, Proceedings, Part II 13}, pages 268--283. Springer, 2014.

\bibitem[Mildenhall et~al.(2021)Mildenhall, Srinivasan, Tancik, Barron, Ramamoorthi, and Ng]{nerf}
Ben Mildenhall, Pratul~P Srinivasan, Matthew Tancik, Jonathan~T Barron, Ravi Ramamoorthi, and Ren Ng.
\newblock Nerf: Representing scenes as neural radiance fields for view synthesis.
\newblock \emph{Communications of the ACM}, 65\penalty0 (1):\penalty0 99--106, 2021.

\bibitem[Moreau et~al.(2022)Moreau, Piasco, Tsishkou, Stanciulescu, and de~La~Fortelle]{lens}
Arthur Moreau, Nathan Piasco, Dzmitry Tsishkou, Bogdan Stanciulescu, and Arnaud de La~Fortelle.
\newblock Lens: Localization enhanced by nerf synthesis.
\newblock In \emph{Conference on Robot Learning}, pages 1347--1356. PMLR, 2022.

\bibitem[Oechsle et~al.(2021)Oechsle, Peng, and Geiger]{unisurf}
Michael Oechsle, Songyou Peng, and Andreas Geiger.
\newblock Unisurf: Unifying neural implicit surfaces and radiance fields for multi-view reconstruction.
\newblock In \emph{Proceedings of the IEEE/CVF International Conference on Computer Vision}, pages 5589--5599, 2021.

\bibitem[Park et~al.(2019)Park, Florence, Straub, Newcombe, and Lovegrove]{deepsdf}
Jeong~Joon Park, Peter Florence, Julian Straub, Richard Newcombe, and Steven Lovegrove.
\newblock Deepsdf: Learning continuous signed distance functions for shape representation.
\newblock In \emph{Proceedings of the IEEE/CVF conference on computer vision and pattern recognition}, pages 165--174, 2019.

\bibitem[Reizenstein et~al.(2021)Reizenstein, Shapovalov, Henzler, Sbordone, Labatut, and Novotny]{common}
Jeremy Reizenstein, Roman Shapovalov, Philipp Henzler, Luca Sbordone, Patrick Labatut, and David Novotny.
\newblock Common objects in 3d: Large-scale learning and evaluation of real-life 3d category reconstruction.
\newblock In \emph{Proceedings of the IEEE/CVF International Conference on Computer Vision}, pages 10901--10911, 2021.

\bibitem[Ren et~al.(2023)Ren, Zhang, Pollefeys, S{\"u}sstrunk, and Wang]{volrecon}
Yufan Ren, Tong Zhang, Marc Pollefeys, Sabine S{\"u}sstrunk, and Fangjinhua Wang.
\newblock Volrecon: Volume rendering of signed ray distance functions for generalizable multi-view reconstruction.
\newblock In \emph{Proceedings of the IEEE/CVF Conference on Computer Vision and Pattern Recognition}, pages 16685--16695, 2023.

\bibitem[Schonberger and Frahm(2016)]{sfm16}
Johannes~L Schonberger and Jan-Michael Frahm.
\newblock Structure-from-motion revisited.
\newblock In \emph{Proceedings of the IEEE conference on computer vision and pattern recognition}, pages 4104--4113, 2016.

\bibitem[Sch{\"o}nberger et~al.(2016)Sch{\"o}nberger, Zheng, Frahm, and Pollefeys]{colmap}
Johannes~L Sch{\"o}nberger, Enliang Zheng, Jan-Michael Frahm, and Marc Pollefeys.
\newblock Pixelwise view selection for unstructured multi-view stereo.
\newblock In \emph{Computer Vision--ECCV 2016: 14th European Conference, Amsterdam, The Netherlands, October 11-14, 2016, Proceedings, Part III 14}, pages 501--518. Springer, 2016.

\bibitem[Sinha et~al.(2023)Sinha, Zhang, Tagliasacchi, Gilitschenski, and Lindell]{sparsepose}
Samarth Sinha, Jason~Y Zhang, Andrea Tagliasacchi, Igor Gilitschenski, and David~B Lindell.
\newblock Sparsepose: Sparse-view camera pose regression and refinement.
\newblock In \emph{Proceedings of the IEEE/CVF Conference on Computer Vision and Pattern Recognition}, pages 21349--21359, 2023.

\bibitem[Truong et~al.(2023)Truong, Rakotosaona, Manhardt, and Tombari]{sparf}
Prune Truong, Marie-Julie Rakotosaona, Fabian Manhardt, and Federico Tombari.
\newblock Sparf: Neural radiance fields from sparse and noisy poses.
\newblock In \emph{Proceedings of the IEEE/CVF Conference on Computer Vision and Pattern Recognition}, pages 4190--4200, 2023.

\bibitem[Ullman(1979)]{sfm79}
Shimon Ullman.
\newblock The interpretation of structure from motion.
\newblock \emph{Proceedings of the Royal Society of London. Series B. Biological Sciences}, 203\penalty0 (1153):\penalty0 405--426, 1979.

\bibitem[Vaswani et~al.(2017)Vaswani, Shazeer, Parmar, Uszkoreit, Jones, Gomez, Kaiser, and Polosukhin]{attention}
Ashish Vaswani, Noam Shazeer, Niki Parmar, Jakob Uszkoreit, Llion Jones, Aidan~N Gomez, {\L}ukasz Kaiser, and Illia Polosukhin.
\newblock Attention is all you need.
\newblock \emph{Advances in neural information processing systems}, 30, 2017.

\bibitem[Wang et~al.(2021{\natexlab{a}})Wang, Liu, Liu, Theobalt, Komura, and Wang]{neus}
Peng Wang, Lingjie Liu, Yuan Liu, Christian Theobalt, Taku Komura, and Wenping Wang.
\newblock Neus: Learning neural implicit surfaces by volume rendering for multi-view reconstruction.
\newblock \emph{arXiv preprint arXiv:2106.10689}, 2021{\natexlab{a}}.

\bibitem[Wang et~al.(2021{\natexlab{b}})Wang, Wang, Genova, Srinivasan, Zhou, Barron, Martin-Brualla, Snavely, and Funkhouser]{ibrnet}
Qianqian Wang, Zhicheng Wang, Kyle Genova, Pratul~P Srinivasan, Howard Zhou, Jonathan~T Barron, Ricardo Martin-Brualla, Noah Snavely, and Thomas Funkhouser.
\newblock Ibrnet: Learning multi-view image-based rendering.
\newblock In \emph{Proceedings of the IEEE/CVF Conference on Computer Vision and Pattern Recognition}, pages 4690--4699, 2021{\natexlab{b}}.

\bibitem[Wang et~al.(2022)Wang, Zhu, Huang, Qin, Ye, He, Chi, and Wang]{mvster}
Xiaofeng Wang, Zheng Zhu, Guan Huang, Fangbo Qin, Yun Ye, Yijia He, Xu Chi, and Xingang Wang.
\newblock Mvster: Epipolar transformer for efficient multi-view stereo.
\newblock In \emph{European Conference on Computer Vision}, pages 573--591. Springer, 2022.

\bibitem[Xu et~al.(2022)Xu, Zhang, Cai, Rezatofighi, and Tao]{gmflow}
Haofei Xu, Jing Zhang, Jianfei Cai, Hamid Rezatofighi, and Dacheng Tao.
\newblock Gmflow: Learning optical flow via global matching.
\newblock In \emph{Proceedings of the IEEE/CVF conference on computer vision and pattern recognition}, pages 8121--8130, 2022.

\bibitem[Xu et~al.(2023{\natexlab{a}})Xu, Zhang, Cai, Rezatofighi, Yu, Tao, and Geiger]{unifyingflow}
Haofei Xu, Jing Zhang, Jianfei Cai, Hamid Rezatofighi, Fisher Yu, Dacheng Tao, and Andreas Geiger.
\newblock Unifying flow, stereo and depth estimation.
\newblock \emph{IEEE Transactions on Pattern Analysis and Machine Intelligence}, 2023{\natexlab{a}}.

\bibitem[Xu et~al.(2023{\natexlab{b}})Xu, Guan, Wang, Liu, Zeng, Wang, and Yang]{c2f2neus}
Luoyuan Xu, Tao Guan, Yuesong Wang, Wenkai Liu, Zhaojie Zeng, Junle Wang, and Wei Yang.
\newblock C2f2neus: Cascade cost frustum fusion for high fidelity and generalizable neural surface reconstruction.
\newblock \emph{arXiv preprint arXiv:2306.10003}, 2023{\natexlab{b}}.

\bibitem[Yao et~al.(2018)Yao, Luo, Li, Fang, and Quan]{mvsnet}
Yao Yao, Zixin Luo, Shiwei Li, Tian Fang, and Long Quan.
\newblock Mvsnet: Depth inference for unstructured multi-view stereo.
\newblock In \emph{Proceedings of the European conference on computer vision (ECCV)}, pages 767--783, 2018.

\bibitem[Yariv et~al.(2020)Yariv, Kasten, Moran, Galun, Atzmon, Ronen, and Lipman]{idr}
Lior Yariv, Yoni Kasten, Dror Moran, Meirav Galun, Matan Atzmon, Basri Ronen, and Yaron Lipman.
\newblock Multiview neural surface reconstruction by disentangling geometry and appearance.
\newblock \emph{Advances in Neural Information Processing Systems}, 33:\penalty0 2492--2502, 2020.

\bibitem[Yariv et~al.(2021)Yariv, Gu, Kasten, and Lipman]{volsdf}
Lior Yariv, Jiatao Gu, Yoni Kasten, and Yaron Lipman.
\newblock Volume rendering of neural implicit surfaces.
\newblock \emph{Advances in Neural Information Processing Systems}, 34:\penalty0 4805--4815, 2021.

\bibitem[Yu et~al.(2021)Yu, Ye, Tancik, and Kanazawa]{pixelnerf}
Alex Yu, Vickie Ye, Matthew Tancik, and Angjoo Kanazawa.
\newblock pixelnerf: Neural radiance fields from one or few images.
\newblock In \emph{Proceedings of the IEEE/CVF Conference on Computer Vision and Pattern Recognition}, pages 4578--4587, 2021.

\bibitem[Yu et~al.(2022)Yu, Peng, Niemeyer, Sattler, and Geiger]{monosdf}
Zehao Yu, Songyou Peng, Michael Niemeyer, Torsten Sattler, and Andreas Geiger.
\newblock Monosdf: Exploring monocular geometric cues for neural implicit surface reconstruction.
\newblock \emph{Advances in neural information processing systems}, 35:\penalty0 25018--25032, 2022.

\bibitem[Zhu et~al.(2022)Zhu, Peng, Larsson, Xu, Bao, Cui, Oswald, and Pollefeys]{niceslam}
Zihan Zhu, Songyou Peng, Viktor Larsson, Weiwei Xu, Hujun Bao, Zhaopeng Cui, Martin~R Oswald, and Marc Pollefeys.
\newblock Nice-slam: Neural implicit scalable encoding for slam.
\newblock In \emph{Proceedings of the IEEE/CVF Conference on Computer Vision and Pattern Recognition}, pages 12786--12796, 2022.

\end{thebibliography}
}

\maketitlesupplementary
\renewcommand\thesection{\Alph{section}}
\renewcommand\thesubsection{\thesection.\arabic{subsection}}
\setcounter{section}{0}

\section{More Implementation Details}
\vspace{0.5ex}\noindent
In this section, we provide more details on our model architectures not fully covered in the main paper.

\vspace{0.5ex}\noindent
\textbf{Network details.}
Given multi-view images, we first use the Feature Pyramid Network~\cite{fpn} to extract image features.
We set the number of channels of each level to 32, 16, and 8, respectively.
Similar to TransMVSNet~\cite{transmvsnet}, we utilize a sequence of attention blocks to extract cross-view features.
To build correlation frustums from each view, every source image is considered as a reference image, while the other images are considered as the source images.
We use 4 attention blocks and we use multi-head attention with the number of heads of 8.

\noindent
\textbf{Building Correlation Frustums.}
We build a cascaded correlation frustum for each source view as Sec~\ref{cross-view volume}. 
We clarify and elaborate more on the aggregation function $\langle \cdot \rangle$ of the correlation frustums here.
For each level $l=1,...,L$, we obtain pairwise feature correlation as follows:
\begin{equation}
    c^{(d)}_{ij}(\textbf{p}) = F_i \cdot \hat{F}^{(d)}_{j \rightarrow i},
\end{equation}
\noindent
where $\cdot$ denotes a dot product and $\hat{F}^{(d)}_{j \rightarrow i}$ denotes a warped $j$-th source feature map to $i$-th source feature at depth hypothesis $d$. 
Then we add all $N-1$ possible pairs to assign a pixel-wise weight map with its maximum correlation along the depth dimension as follows:
\begin{equation}
    C^{(d)}_i(\textbf{p}) = \sum_{\substack{j=1 \\ j \neq i}}^{N} \max_{d}\left\{ c^{(d)}_{ij}(\textbf{p}) \right\} \cdot c^{(d)}_{ij}(\textbf{p}).
\end{equation}
Finally, we concatenate all depth hypotheses and add correlation frustums of all source viewpoints:
\begin{equation}
    C^{\prime}(\textbf{p}) = \sum_{i=1}^{N}{C}_i(\textbf{p}),
\end{equation}
where $C^{\prime}(\textbf{p}) \in \mathcal{R}^{1 \times d \times h \times w}$ denotes the combined correlation frustums from all source views. 
This combined correlation frustum is regularized with 3D CNNs and output $V_i \in \mathcal{R}^{c \times d \times h \times w}$, which is used for estimating intermediate depths.
We further regularize $V_i$ to represent global feature volume $V^{\prime}_i$ and we obtain the global feature vector via trilinear interpolation for each level $l$.

\begin{table*}[t!]
\centering
\small
\resizebox{\textwidth}{!}{ 
\centering
\small
\begin{tabular}{l|lcccccccccccccccc}
\hline
Set       & Method & 24 & 37 & 40 & 55 & 63 & 65 & 69 & 83 & 97 & 105 & 106 & 110 & 114 & 118 & 122 & Mean (CD) $\downarrow$ \\
\hline \hline
            & VolRecon~\cite{volrecon} & 1.20 & 2.59 & 1.56 & 1.08 & 1.43 & 1.92 & 1.11 & 1.48 & 1.42 & 1.05 & 1.19 & 1.38 & 0.74 & 1.23 & 1.27 &  1.38 \\
Favorable   & ReTR~\cite{retr}         & 1.05 & 2.31 & 1.44 & 0.98 & 1.18 & 1.52 & 0.88 & 1.35 & 1.30 & 0.87 & 1.07 & 0.77 & 0.59 & 1.05 & 1.12 &  1.17 \\
(VC: 1788)  & Ours                     & \textbf{0.76} & \textbf{2.05} & \textbf{1.31} & \textbf{0.82} & \textbf{1.12} & \underline{1.18} & \underline{0.74} & \textbf{1.17} & \textbf{1.11} & \textbf{0.71}                                             & \textbf{0.88} & \underline{0.58} & \underline{0.54} & \textbf{0.86} & \underline{0.99} &  \textbf{0.99} \\
            & Ours*                    & \underline{0.77} & \underline{2.10} & \underline{1.34} & \underline{0.87} & \underline{1.15} & \textbf{1.16} & \textbf{0.71} & \underline{1.25} & \underline{1.17} & \underline{0.81} & \underline{0.90} & \textbf{0.57} & \textbf{0.51} & \textbf{0.86} & \textbf{0.97} & \underline{1.01} \\
\hline
            & VolRecon~\cite{volrecon} & 2.63 & 4.22 & 2.89 & 2.49 & 2.93 & 2.50 & 1.68 & 1.84 & 2.02 & 1.76 & 2.35 & 2.64 & 1.16 & 2.17 & 1.76 & 2.34 \\
Normal      & ReTR~\cite{retr}         & 2.06 & 3.72 & 2.54 & 2.51 & 1.75 & 2.11 & 1.49 & 1.57 & 1.74 & 1.35 & 1.88 & 2.05 & 1.00 & 1.74 & 1.48 & 1.93 \\
(VC: 192)   & Ours                    & \underline{1.30} & \underline{2.59} & \underline{1.51} & \underline{1.39} & \textbf{1.04} & \underline{1.28} & \underline{0.80} &           
                                         \underline{1.37} & \underline{1.16} & \underline{0.95} & \underline{0.98} & \underline{0.90} & \underline{0.54} & \underline{1.06} & \underline{1.08} & \underline{1.20} \\
            & Ours*                      & \textbf{1.02} & \textbf{2.21} & \textbf{1.42} & \textbf{1.00} & \underline{1.23} & \textbf{1.24} & \textbf{0.72} & \textbf{1.36} & \textbf{1.03} &                                           \textbf{0.77} & \textbf{0.86} & \textbf{0.84} & \textbf{0.47} & \textbf{0.91} & \textbf{0.96} & \textbf{1.07} \\
                                         \hline
                                         
            & VolRecon~\cite{volrecon} & 3.43 & 3.64 & 4.26 & 4.63 & 2.43 & 3.40 & 2.81 & 2.41 & 2.36 & 2.49 & 3.79 & 3.55 & 1.44 & 3.60 & 3.38 &  3.18 \\
Unfavorable & ReTR~\cite{retr}         & 3.00 & 3.98 & 3.78 & 4.22 & 2.22 & 2.93 & 3.00 & 2.51 & 2.24 & 2.36 & 2.36 & 3.92 & 1.63 & 2.83 & 3.07 &  2.94 \\
(VC: 57)    & Ours                     & \underline{1.39} & \underline{2.25} & \underline{1.65} & \underline{1.96} & \underline{1.53} & \underline{1.61} & \underline{1.22} &           
                                  \underline{1.92} & \underline{1.36} & \underline{1.66} & \underline{1.75} & \underline{1.29} & \underline{0.73} & \underline{1.70} & \underline{1.39} &  \underline{1.56} \\
            & Ours*                    & \textbf{1.31} & \textbf{2.00} & \textbf{1.41} & \textbf{1.36} & \textbf{1.24} & \textbf{1.58} & \textbf{1.06} & \textbf{1.44} & \textbf{1.37} & 
                                         \textbf{0.99} & \textbf{1.45} & \textbf{0.96} & \textbf{0.58} & \textbf{1.34} & \textbf{1.09} &  \textbf{1.28} \\
\hline
\end{tabular}
}
\caption{\textbf{Quantitative results on different View-Combination (VC) score levels.} Each method's performance is indicated across all scans, with the mean Chamfer Distance (CD). An asterisk (*) indicates methods trained with the random set training strategy (Sec.\ref{Training}). We use camera indices (23,24,33), (16,26,42), and (1,16,36) for \textit{Favorable}, \textit{Normal}, \textit{Unfavorable} sets respectively. The corresponding average \textit{VC} scores are given in parentheses.}
\label{table:group_comparison}
\end{table*}

\section{Detailed Description of Baseline Methods}
\vspace{0.5ex}\noindent
In our research, we compared our method with four different types of surface reconstruction approaches across all experiments.
(1) The most directly comparable category is generalizable implicit surface reconstruction~\cite{sparseneus,volrecon,retr,c2f2neus}, where our method is included.
These approaches assume a very limited number of views and unseen scenes during inference.
(2) Generalizable neural rendering methods~\cite{pixelnerf,ibrnet,mvsnerf} also use a very limited number of views to perform novel view synthesis from unseen scenes.
These methods learn geometry through volume density rather than SDF, leading to ambiguity as they lack an explicit surface definition. By applying thresholding on volume density and with Marching Cubes~\cite{marchingcubes}, geometry can be obtained.
(3) Neural implicit surface reconstruction~\cite{volsdf,neus} techniques extract accurate geometry from densely captured multi-view images of a single scene via optimization. Since these methods cannot generalize across multiple scenes, per-scene optimization is required to estimate geometry, unlike generalizable methods.
(4) We compare with a conventional matching-based method, COLMAP~\cite{colmap}. It utilizes feature matching and triangulation from multiple images, providing a baseline for comparing the effectiveness of neural network-based methods.
(5) Additionally, we include learning-based MVS methods~\cite{mvsnet,transmvsnet}. 

We only use generalizable implicit surface reconstruction methods for comparing view-combination generalizability in Table~\ref{table:unfav_comparison} as they outperform all the other approaches in terms of generalizable surface reconstruction.

\section{More Information on the VC Score}
\vspace{0.5ex}\noindent
\textbf{Derivation from view selection score.}
In this section, we provide more details on the View-Combination (\textit{VC}) score which is derived from the view selection score~\cite{mvsnet} as described in Sec.~\ref{sec:motivation}.
View selection score $s(i,j)$ indicates the informativeness of a view pair between image $i$ and image $j$.
\begin{equation}
\begin{aligned}
s(i,j) &= \sum_\mathbf{p}\mathcal{G}(\theta_{ij}(\mathbf{p})), \\
\end{aligned}
\end{equation}
\begin{equation}
\begin{aligned}
\theta_{ij}(\mathbf{p}) &= \left(\frac{180}{\pi}\right)\arccos{((c_i - \mathbf{p}) \cdot (c_j - \mathbf{p}))},
\end{aligned}
\end{equation}
where $\textbf{p}$ is a common track in both view $i$ and $j$, while $c_i$ and $c_j$ denote the camera center of view $i$ and view $j$, respectively.
To obtain $\textbf{p}$, we use off-the-shelf reconstruction software, COLMAP~\cite{colmap}. 
In addition, $\mathcal{G}$ is a piecewise Gaussian function that peaks at certain angle $\theta_{0}$ as follows:
\begin{equation}
\mathcal{G}(\theta) = \begin{cases}
\exp{(-\frac{(\theta-\theta_0)^2}{2\sigma_1^2})}, & \theta \leq \theta_0, \\
\exp{(-\frac{(\theta-\theta_0)^2}{2\sigma_2^2})}, & \theta > \theta_0,
\end{cases}
\end{equation}
where we set $\theta_0=5$, $\sigma_1 = 1$, and $\sigma_2 = 10$ following the original definition~\cite{mvsnet}.
While the view selection score originally calculates between two image pair, we define \textit{VC} score to represent a collective informativeness of a combination for the reconstruction by averaging the values for all possible pairs as follows:
\begin{equation}
\textit{VC} = \frac{1}{\binom{n}{2}} \sum_{i=1}^{n-1} \sum_{j=i+1}^{n} s(i,j),
\end{equation}
where $n$ denotes the number of source images.

\begin{figure}[h] 
    \centering
    \includegraphics[width=1.0\linewidth, trim={0cm 0cm 0cm 0cm}, clip]{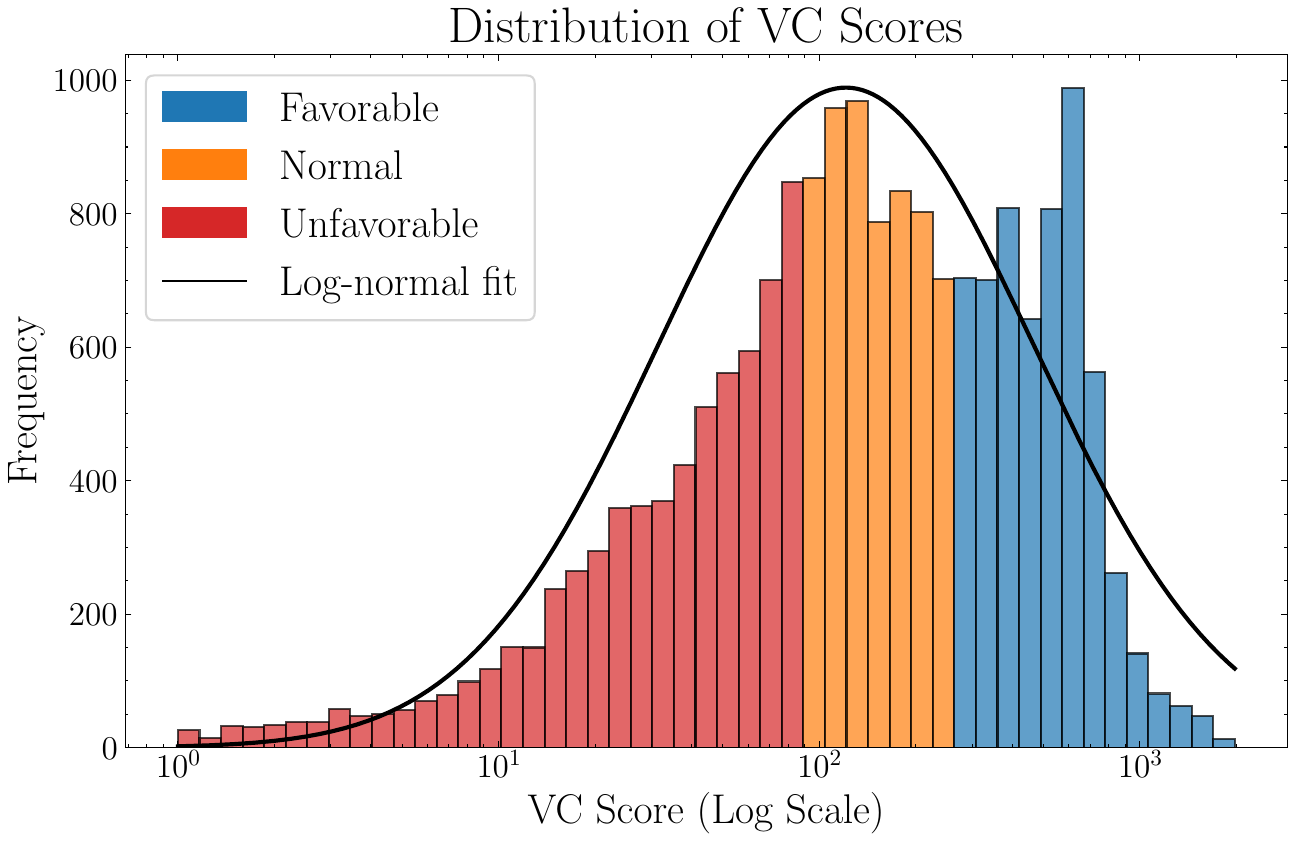}
    \caption{\textbf{Distribution of the View-Combination (\textit{VC}) scores on the test scenes}. We visualize the statistical distribution of all scenes to give a sense of how \textit{VC} score is distributed. We average all the scores of the test scenes given a camera view combination. A curved line denotes a fitted Gaussian distribution to VC scores in the log scale.}
    \label{fig:dist}
\end{figure}

\vspace{0.5ex}
\noindent
\textbf{VC score statistics.}
\vspace{0.5ex}\noindent
For more statistical information about the \textit{VC} score, we report the distribution of the \textit{VC} score on the DTU test dataset that we used for the experiments.
We average the \textit{VC} scores of all test scenes for each combination. As shown in Figure~\ref{fig:dist}, we evenly split the entire scores into three groups, \textit{Favorable, Normal} and \textit{Unfavorable}. The distribution is bell-shaped where the \textit{Normal} set generally has the highest frequencies.     
To reproduce the Table~\ref{table:fav_comparison} and Table~\ref{table:unfav_comparison}, we use the combination with the camera indices $(23, 24, 33)$ as a favorable set following the previous generalizable surface reconstruction protocols~\cite{sparseneus,volrecon,retr,c2f2neus}, and $(1, 16, 36)$ for the unfavorable set.

\section{Results on Various VC Score Levels}
We have reported the reconstruction performance on both \textit{Unfavorable} (Table~\ref{table:unfav_comparison}) and \textit{Favorable} (Table~\ref{table:fav_comparison}) sets. Additionally, we provide a comprehensive report on performance across all \textit{VC} score levels, which includes results for the \textit{Normal} set. It is important to note that, unless specified otherwise, we adhered to the same training protocol as in previous studies~\cite{volrecon,retr}, utilizing only fixed sets of the best view combinations.

As demonstrated in Table~\ref{table:group_comparison}, our method consistently shows superior performance across all VC levels and scenes. Furthermore, when employing the random set strategy for training (denoted as Ours* in Table~\ref{table:group_comparison}), we observed an enhanced performance, particularly in the \textit{Normal} and \textit{Unfavorable} sets. This finding confirms the effectiveness of the random set training strategy in improving generalizability across different view combinations.

\section{Random Set Training on Baseline Methods}

The random set training strategy, intuitively, should be beneficial in reconstructing from \textit{Unfavorable} sets, as it exposes the network to a variety of view combinations during training. Meanwhile, this approach might slightly degrade the performance in \textit{Favorable} sets compared to methods trained solely with fixed \textit{Favorable} sets.

To assess the impact of random set training, we applied this strategy to baseline methods~\cite{volrecon, retr} and reported the results in Table~\ref{tab:random_set_baseline}. In the case of VolRecon~\cite{volrecon}, random set training resulted in suboptimal performance in both \textit{Favorable} and \textit{Unfavorable} test scenarios. We conjecture that the introduction of arbitrary view combinations as inputs may disrupt the training dynamics in VolRecon~\cite{volrecon}.

In contrast, for both ReTR~\cite{retr} and our method, the adoption of random set training improved performance in \textit{Unfavorable} scenarios, albeit at the expense of slightly diminished results in \textit{Favorable} sets. Notably, our approach with random set training (Ours*) exhibited a minor performance degradation in \textit{Favorable} sets (\textcolor{red}{-2\%}), while achieving a significant enhancement in \textit{Unfavorable} sets (\textcolor{blue}{+18\%}). This improvement can be attributed to our framework's ability to learn correlations among source images.

\begin{table}
    \centering
    \begin{tabular}{ccc}
    \hline
        Method & Favorable & Unfavorable \\ \hline \hline
        VolRecon~\cite{volrecon} & 1.42 & 3.18 \\
        VolRecon*~\cite{volrecon} & 2.74~\textcolor{red}{(+1.32, 92\%)} & 3.88~\textcolor{red}{(+0.7, 22\%)}  \\ \hline
        ReTR~\cite{retr} & 1.17 & 2.94 \\
        ReTR*~\cite{retr} & 1.62~\textcolor{red}{(+0.45, 38\%)} & 2.88~\textcolor{blue}{(-0.06, 2\%)} \\ \hline
        Ours & 0.99 & 1.56 \\
        Ours* & 1.01~\textcolor{red}{(+0.02, 2\%)} & 1.28 \textcolor{blue}{(-0.28, 18\%)} \\
    \hline
    \end{tabular}
    \caption{\textbf{Effect of employing the random set training.} This table shows the Chamfer Distance (CD) metrics across different methods, comparing the performance on both \textit{Favorable} and \textit{Unfavorable} sets. The lower CD represents the better results. (*) denote the use of a random set training strategy (Sec.~\ref{Training}). The values in parentheses represent the change in performance relative to the standard training approach, with percentages indicating the rate of degradation (\textcolor{red}{+}) or improvement (\textcolor{blue}{-}).}
    \label{tab:random_set_baseline}
\end{table}

\section{Impact of the Number of Views}
We explored the effect of viewpoint density on our method by varying the number of source views. As detailed in Table~\ref{tab:view_num}, we observed a gradual improvement in performance with increasing viewpoint density. Using more views alleviates the challenges associated with reconstructing difficult regions, such as those that are occluded or not commonly shared across views.

\begin{table}[t!]
\centering
\begin{tabular}{ccc}
\hline
Number of Views & VolRecon~\cite{volrecon} & Ours \\
\hline
2 & 1.72 & 1.15 \\
3 & 1.38 & 1.00 \\
4 & 1.35 & 0.97 \\
5 & \textbf{1.33} & \textbf{0.96} \\
\hline
\end{tabular}
\caption{\textbf{Effect of the number of views on reconstruction.} The table compares the Chamfer Distance for VolRecon~\cite{volrecon} and our method across varying number of viewpoints. As the number of views increases, the reconstruction quality improves for both methods, with our method showing superior performance in all tested cases.}
\label{tab:view_num}
\end{table}

\section{Comparison on Depth Estimation.}
MVS methods and other baseline methods experience significant performance degradation in unfavorable settings (Table~\ref{tab:depth_comparison}), which are more susceptible to self-occlusion. The correlation frustum is one of our key contributions (not introduced in C2F2NeuS) to model global correlation among input views, and the reconstruction transformer estimates SDFs in a correlation-aware manner. We achieve robust performance in handling arbitrary view combinations.

\begin{table}[h]
\centering
\scriptsize
\setlength{\tabcolsep}{4.0pt} 
\renewcommand{\arraystretch}{1.0}
\begin{tabular}{p{1.4cm}cccccc}
\toprule
\multicolumn{1}{c}{} & \multicolumn{3}{c}{Favorable} & \multicolumn{3}{c}{Unfavorable} \\
\cmidrule(lr){2-4} \cmidrule(lr){5-7}
Method & \textless 1mm $\uparrow$ & \textless 4mm $\uparrow$ & Abs. $\downarrow$ & \textless 1mm $\uparrow$ & \textless 4mm $\uparrow$ & Abs. $\downarrow$ \\
\midrule
MVSNet [43] & 29.95 & 72.33 & 13.62 & 5.82 & 19.72 & 72.74 \\
TransMVSNet[8] & 38.95 & 82.91 & 13.53 & 6.21 & 15.60 & 161.71 \\
VolRecon & 43.60 & 82.53 & 7.12 & 3.93 & 11.43 & 185.95 \\
ReTR [22] & 51.95 & 85.06 & \textbf{5.41} & 10.83 & 19.26 & 58.02 \\
Ours & \textbf{54.80} & \textbf{86.69} & 6.09 & \textbf{51.30} & \textbf{65.14} & \textbf{17.02} \\
\bottomrule
\end{tabular}
\caption{3-Views Depth Map Comparison on DTU Datasets. 
}
\label{tab:depth_comparison}
\end{table}
 

\section{Comparative Ablation Study on Depth Supervision}
As shown in Table~\ref{tab:depth_ablation}, our method works well even without depth supervision.
Especially in unfavorable sets, our method without depth supervision outperforms the baselines even with depth supervision. Depth supervision in SparseNeuS tends to result in over-smooth surfaces, which is also reported in~\cite{volrecon}.

\begin{table}[h]
\centering
\scriptsize
\setlength{\tabcolsep}{10pt} 
\begin{tabular}{p{1.8cm}cccc}
\toprule
\multicolumn{1}{c}{} & \multicolumn{2}{c}{Favorable (CD$\downarrow$)} & \multicolumn{2}{c}{Unfavorable (CD$\downarrow$)} \\
\cmidrule(lr){2-3} \cmidrule(lr){4-5}
Method & w/o $\mathcal{L}_d$ & w/ $\mathcal{L}_d$ & w/o $\mathcal{L}_d$ & w/ $\mathcal{L}_d$ \\
\midrule
SparseNeuS [24] & 1.64 & 4.22 & 4.16 & 5.55 \\
VolRecon [30]   & 2.06 & 1.38 & 8.65 & 3.18 \\
ReTR [22]       & 1.45 & 1.17 & 4.41 & 2.94 \\
Ours       & \textbf{1.26} & \textbf{0.99} & \textbf{2.26} & \textbf{1.56} \\
\bottomrule
\end{tabular}
\caption{Ablation study of Depth Supervision ($\mathcal{L}_d$).}
\label{tab:depth_ablation}
\end{table}

\section{Regarding Pracitcal Relevance.} 
Apparently, obtaining accurate camera poses from a sparse set of images presents challenges. 
In response, a parallel line of research has focused on estimating relative camera poses from sparse views~\cite{sparsepose,idpose}, 
and it is anticipated that we can apply these methods to obtain more accurate camera poses from sparse, unfavorable set of views.
Another research direction explores reconstructing 3D in a pose-free manner only using source images~\cite{posefreegen,leap}, in which case our proposed modeling of correlation across images becomes even more meaningful.
These approaches potentially address concerns on obtaining camera poses. 
We believe that obtaining accurate geometry under pose-free settings remains underdeveloped and seems to be a promising future work. 

\section{More Qualitative Results}
\vspace{0.5ex}\noindent
Lastly, we provide additional visual examples that highlight the effectiveness of our method.
We include all test scenes except the scenes reported in the main paper.
Similar to Figure~\ref{fig:qualitative}, we include the results of the comparison with baseline methods across \textit{Favorable}, \textit{Normal}, and \textit{Unfavorable} sets.

\begin{figure*}[t]
    \centering
    \includegraphics[width=0.9\linewidth, trim={0cm 1.50cm 9.75cm 0cm}, clip]{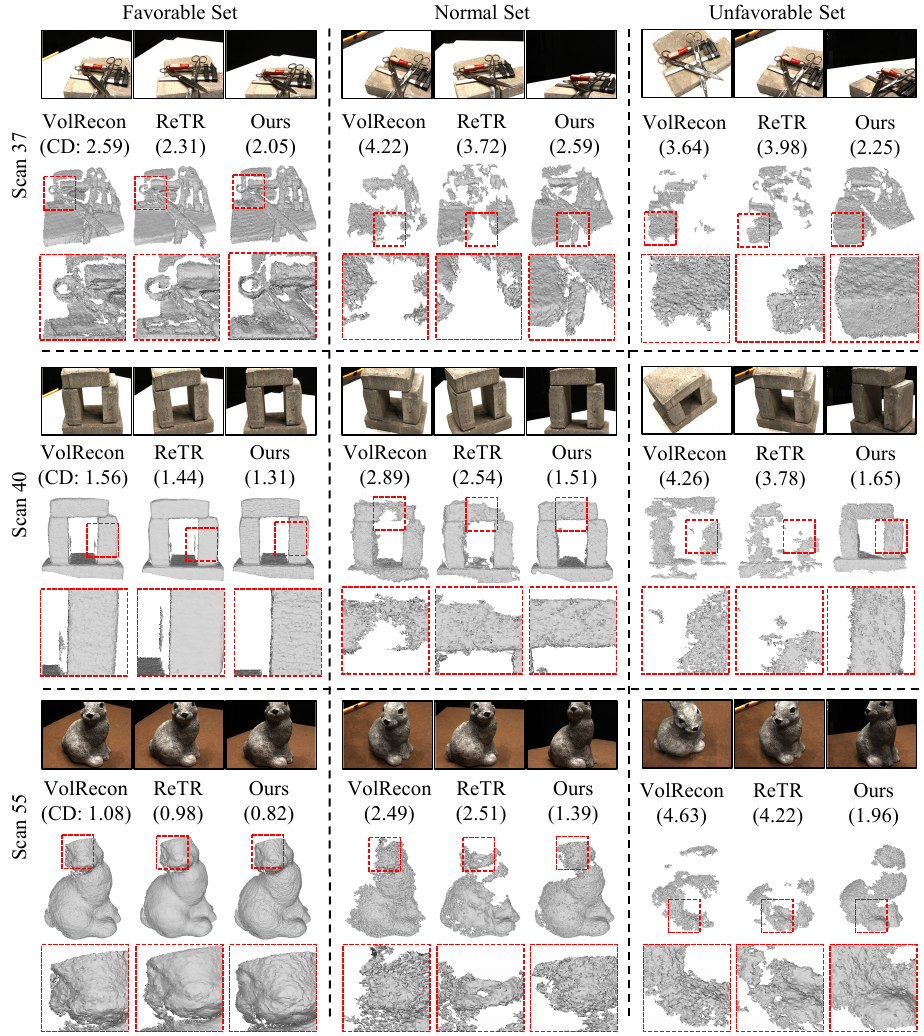}
    \caption{\textbf{A qualitative results of reconstruction across various VC Levels.} The numbers in parentheses denote the Chamfer Distance.}
    \label{fig:supl_qual1}
    \vspace{-2mm}
\end{figure*}
    
\begin{figure*}[t]
    \centering
    \includegraphics[width=0.9\linewidth, trim={0cm 1.8cm 9.8cm 0cm}, clip]{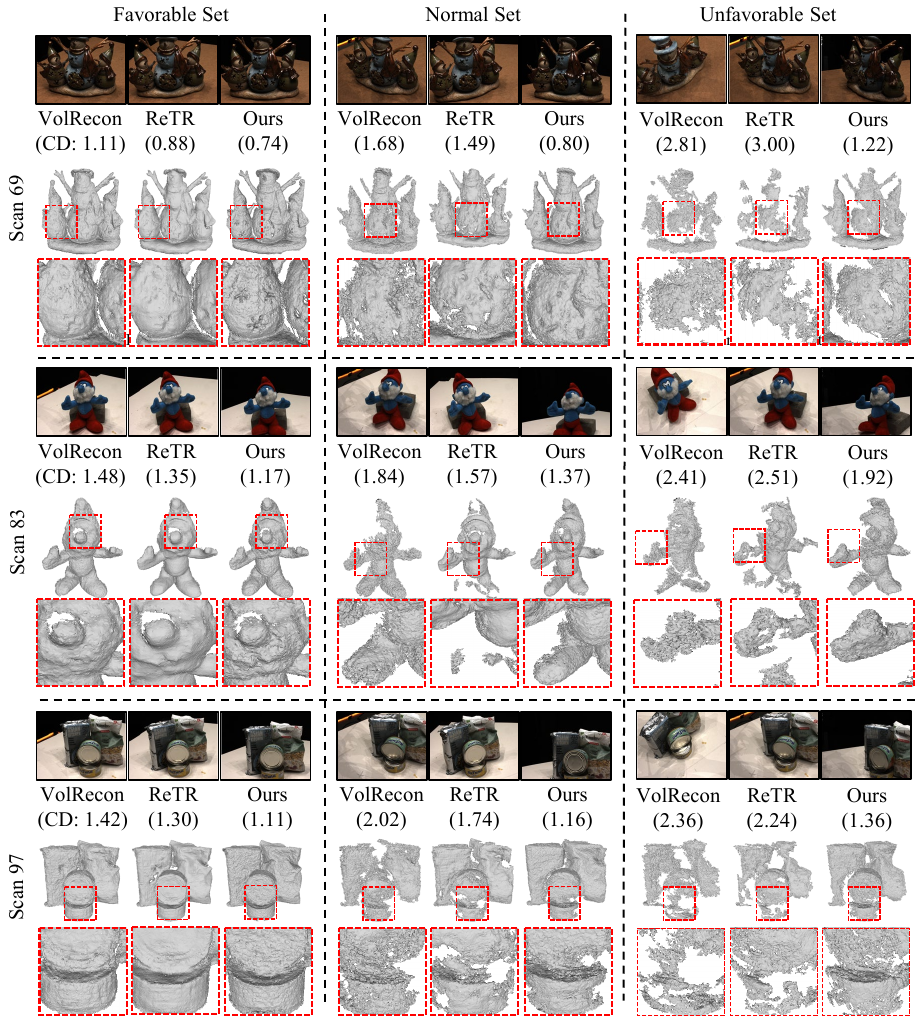}
    \caption{\textbf{A qualitative results of reconstruction across various VC Levels.} The numbers in parentheses denote the Chamfer Distance.}
    \label{fig:supl_qual2}
    \vspace{-2mm}
\end{figure*}

\begin{figure*}[t]
    \centering
    \includegraphics[width=0.9\linewidth, trim={0cm 1.7cm 9.8cm 0cm}, clip]{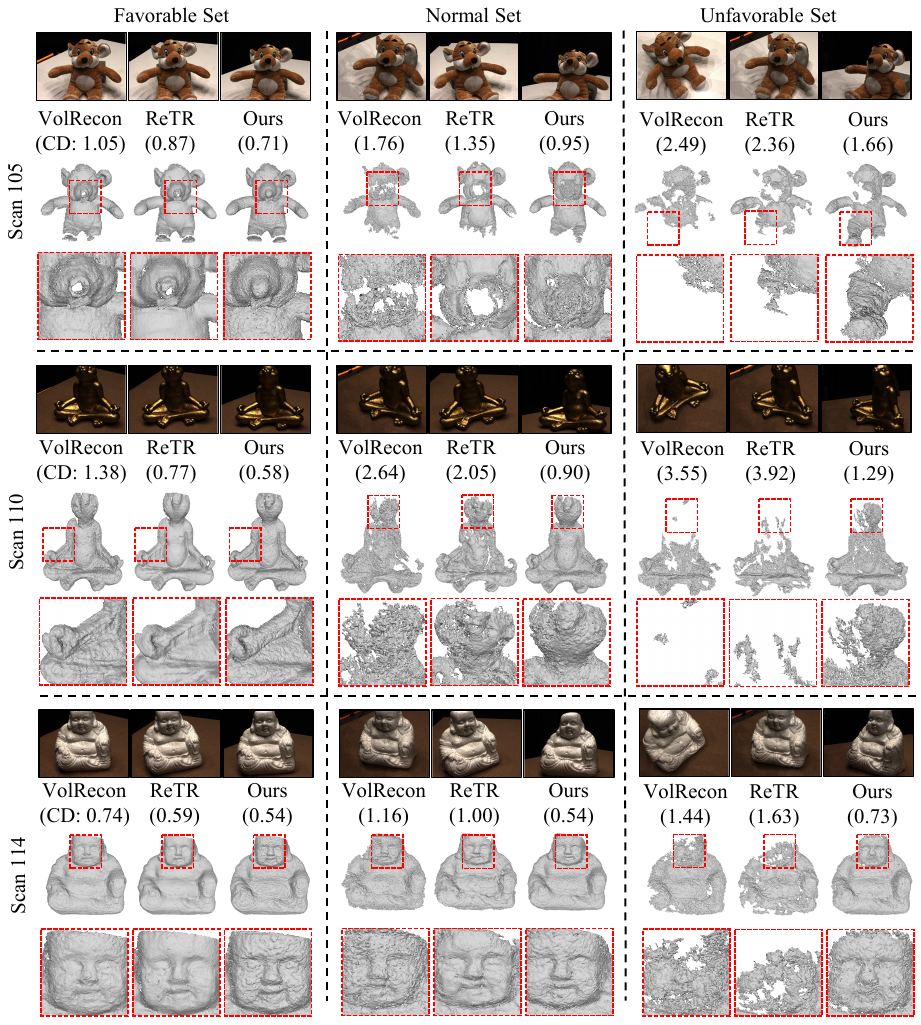}
    \caption{\textbf{A qualitative results of reconstruction across various VC Levels.} The numbers in parentheses denote the Chamfer Distance.}
    \label{fig:supl_qual3}
    \vspace{-2mm}
\end{figure*}

\begin{figure*}[t]
    \centering
    \includegraphics[width=0.9\linewidth, trim={0cm 7.05cm 9.8cm 0cm}, clip]{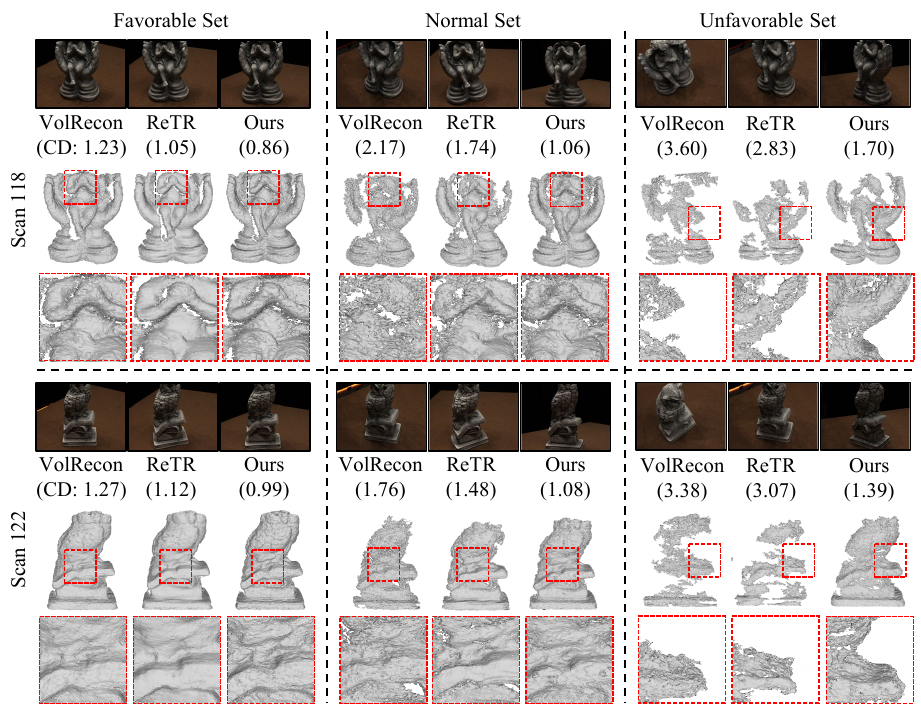}
    \caption{\textbf{A qualitative results of reconstruction across various VC Levels.} The numbers in parentheses denote the Chamfer Distance.}
    \label{fig:supl_qual4}
    \vspace{-2mm}
\end{figure*}



\end{document}